\documentclass[prodmode,acmtkdd]{acmsmall} 

\usepackage[ruled]{algorithm2e}

\SetAlFnt{\small}
\SetAlCapFnt{\small}
\SetAlCapNameFnt{\small}
\SetAlCapHSkip{0pt}
\IncMargin{-\parindent}

\acmVolume{VV}
\acmNumber{NN}
\acmArticle{AA}
\acmYear{YYYY}
\acmMonth{00}

\usepackage{url}
\usepackage{latexsym}
\usepackage{color}
\usepackage[usenames]{xcolor}

\usepackage{amssymb}
\usepackage{graphicx}
\usepackage{rotating}
\usepackage{multirow}
\usepackage{pdfsync}
\usepackage{url}

    \usepackage[applemac]{inputenc}                                          %

  \newif\ifdraft
  \drafttrue

\newcommand{\hidden}[1]{\vspace{0ex}}

\begin{document}

\markboth{Andrea Esuli and Fabrizio Sebastiani}{Optimizing Text Quantifiers for Multivariate Loss Functions}

\title{Optimizing Text Quantifiers for Multivariate Loss Functions}
\author{ANDREA ESULI
\affil{Consiglio Nazionale delle Ricerche}
FABRIZIO SEBASTIANI
\affil{Qatar Computing Research Institute}
}

\begin{abstract}
We address the problem of \emph{quantification}, a
  supervised learning task whose goal is, given a class, to estimate
  the relative frequency (or \emph{prevalence}) of the class in a
  dataset of unlabelled items. Quantification has several applications
  in data and text mining, such as estimating the prevalence of
  positive reviews in a set of reviews of a given product, or
  estimating the prevalence of a given support issue in a dataset of
  transcripts of phone calls to tech support. So far, quantification
  has been addressed by learning a general-purpose classifier, counting the
  unlabelled items which have been assigned the class, and tuning the
  obtained counts according to some heuristics. In this paper we
  depart from the tradition of using general-purpose classifiers, and use
  instead a supervised learning model for \emph{structured
  prediction}, capable of generating classifiers directly optimized
  for the (multivariate and non-linear) function used for evaluating
  quantification accuracy. The experiments that we have run on
  5500 binary high-dimensional datasets 
  (averaging more than
  14,000 documents each)
  show that this method is more accurate, more stable, and more
  efficient than existing, state-of-the-art quantification methods.\end{abstract}



\acmformat{Andrea Esuli and Fabrizio Sebastiani, YYYY. Optimizing Text Quantifiers for Multivariate Loss Functions.}

\begin{bottomstuff}
Authors' address: Andrea Esuli, Istituto di Scienza e Tecnologie
  dell'Informazione, Consiglio Nazionale delle Ricerche, Via Giuseppe
  Moruzzi 1, 56124 Pisa, Italy. E-mail:
  andrea.esuli@isti.cnr.it. Fabrizio Sebastiani, Qatar Computing
  Research Institute, PO Box 5825, Doha, Qatar. E-mail:
  fsebastiani@qf.org.qa. Fabrizio
  Sebastiani is on leave from Consiglio Nazionale delle Ricerche. The order in which the authors are
  listed is purely alphabetical; each author has given an equally
  important contribution to this work.
\end{bottomstuff}

\maketitle


\section{Introduction}
\label{sec:introduction}

\noindent In recent years it has been pointed out that, in a number of
applications involving classification, the final goal is not
determining which class (or classes) individual unlabelled data items
belong to, but determining the \emph{prevalence} (or ``relative
frequency'') of each class in the unlabelled data. The latter task is
known as \emph{quantification}
\cite{Forman:2005fk,Forman06,Forman:2008kx,Forman:2006uq}.

Although what we are going to discuss here applies to any type of
data, we are mostly interested in \emph{text} quantification, i.e.,
quantification when the data items are textual documents. To see the
importance of text quantification, let us examine the task of
classifying textual answers returned to open-ended questions in
questionnaires \cite{Esuli:2010kx,Gamon:2004:SCC,JASIST03}, and let us
discuss two important such scenarios.

In the first scenario, a telecommunications company asks its current
customers the question ``How satisfied are you with our mobile phone
services?'', and wants to classify the resulting textual answers
according to whether they belong to the class
\textsf{MayDefectToCompetition}. The company is likely interested in
accurately classifying each individual customer, since it may want to
call each customer that is assigned the class and offer her improved
conditions.

In the second scenario, a market research agency asks respondents the
question ``What do you think of the recent ad campaign for product
X?'', and wants to classify the resulting textual answers according to
whether they belong to the class \textsf{LovedTheCampaign}. Here, the
agency is likely \emph{not} interested in whether a specific
individual belongs to the class \textsf{LovedTheCampaign}, but is
likely interested in knowing \emph{how many} respondents belong to it,
i.e., in knowing the prevalence of the class.

In sum, while in the first scenario classification is the goal, in the
second scenario the real goal is quantification, i.e., evaluating the
results of classification at the \emph{aggregate} level rather than at
the \emph{individual} level. Other scenarios in which quantification
is the goal may be, e.g., predicting election results by estimating
the prevalence of blog posts (or tweets) supporting a given candidate
or party \cite{Hopkins:2010fk}, or planning the amount of human
resources to allocate to different types of issues in a customer
support center by estimating the prevalence of customer calls related
to a given issue \cite{Forman:2005fk}, or supporting epidemiological
research by estimating the prevalence of medical reports in which a
specific pathology is diagnosed \cite{Baccianella:2013fg}.

The obvious method for dealing with the latter type of scenarios is
\emph{aggregative quantification}, i.e., classifying each
unlabelled document and estimating class prevalence by counting the
documents that have been attributed the class. However, there are two
reasons why this strategy is suboptimal.  The first reason is that a
good classifier may not be a good quantifier, and vice versa. To see
this, one only needs to look at the definition of $F_{1}$, the
standard evaluation function for binary classification, defined as
\begin{equation}\label{eq:f1}F_{1}=\displaystyle\frac{2\cdot
  TP}{2\cdot TP+FP+FN}
\end{equation}
\noindent where $TP$, $FP$ and $FN$ indicate the numbers of true
positives, false positives, and false negatives,
respectively. According to $F_{1}$, a binary classifier $\hat\Phi_{1}$
for which $FP=20$ and $FN=20$ is worse than a classifier
$\hat\Phi_{2}$ for which, on the same test set, $FP=0$ and
$FN=10$. However, $\hat\Phi_{1}$ is intuitively a better binary
quantifier than $\hat\Phi_{2}$; indeed, $\hat\Phi_{1}$ is a perfect
quantifier, since $FP$ and $FN$ are equal and thus compensate each
other, so that the distribution of the test items across the class and
its complement is estimated perfectly.

A second reason is that standard supervised learning algorithms are
based on the assumption that the training set is drawn from the same
distribution as the unlabelled data the classifier is supposed to
classify. But in real-world settings this assumption is often
violated, a phenomenon usually referred to as \emph{concept drift}
\cite{Sammut:2011fk}. For instance, in a backlog of newswire stories
from year 2001, the prevalence of class \textsf{Terrorism} in August
data will likely not be the same as in September data; training on
August data and testing on September data might well yield low
quantification accuracy. Violations of this assumption may occur ``for
reasons ranging from the bias introduced by experimental design, to
the irreproducibility of the testing conditions at training time''
\cite{Quinonero09}. Concept drift usually comes in one of three forms
\cite{Kelly:1999fk}: (a) the class priors $p(c_{i})$ may change, i.e.,
the one in the test set may significantly differ from the one in the
training set; (b) the class-conditional distributions
$p(\mathbf{x}|c_{i})$ may change; (c) the posterior distribution
$p(c_{i}|\mathbf{x})$ may change. It is the first of these three cases
that poses a problem for quantification.

The previous arguments indicate that text quantification should not be
considered a mere byproduct of text classification, and should be
studied as a task of its own. To date, proposed methods explicitly
addressed to quantification (see e.g.,
\cite{Bella:2010kx,Forman:2005fk,Forman06,Forman:2008kx,Forman:2006uq,Hopkins:2010fk,Xue:2009uq})
employ general-purpose supervised learning methods, i.e., address
quantification by elaborating on the results returned by a
general-purpose standard classifier. In this paper we take a sharply
different, \emph{structured prediction} approach, based upon the use
of classifiers explicitly optimized for the non-linear, multivariate
evaluation function that we will use for assessing quantification
accuracy. This idea was first proposed, but not implemented, in
\cite{Esuli:2010fk}.

The rest of the paper is organized as follows. In Section
\ref{sec:preliminaries}, after setting the stage we describe the
evaluation function we will adopt (\S \ref{sec:evaluationmeasures})
and sketch a number of quantification methods previously proposed in
the literature (\S \ref{sec:existingquantificationmethods}). In
Section \ref{sec:novelquantificationmethods} we introduce our novel
method based on explicitly minimizing, via a structured prediction
model, the evaluation measure we have chosen.
Section
\ref{sec:experiments} presents experiments in which we test the method
we propose on two large batches of binary, high-dimensional, publicly
available datasets (the two batches consist of 5148 and 352 datasets,
respectively), using all the methods introduced in \S
\ref{sec:existingquantificationmethods} as baselines.  Section
\ref{sec:related} discusses related work, while Section
\ref{sec:conclusions} concludes.


\section{Preliminaries}
\label{sec:preliminaries}

\noindent In this paper we will focus on quantification at the binary
level. That is, given a domain of documents $D$ and a class $c$, we
assume the existence of an unknown \emph{target function} (or
\emph{ground truth}) $\Phi:D\rightarrow\{-1,+1\}$ that specifies which
members of $D$ belong to $c$; as usual, $+1$ and $-1$ represent
membership and non-membership in $c$, respectively. The approaches we
will focus on are based on aggregative quantification, i.e.,
they rely on the generation of a classifier
$\hat{\Phi}:D\rightarrow\{-1,+1\}$ via supervised learning from a
training set $Tr$. We will indicated with $Te$ the test set on which
quantification effectiveness is going to be tested.



We define the \emph{prevalence} (or \emph{relative frequency})
$\lambda_{Te}(c)$ of class $c$ in a set of documents $Te$ as the
fraction of members of $Te$ that belong to $c$, i.e., as
\begin{equation}\label{eq:prevalence}
  \lambda_{Te}(c)=\frac{|\{d_{j}\in Te|\Phi(d_{j})=+1\}|}{|Te|}
\end{equation}
\noindent Given a set $Te$ of unlabelled documents and a class $c$,
quantification is defined as the task of estimating $\lambda_{Te}(c)$,
i.e., of computing an estimate $\hat{\lambda}_{Te}(c)$ such that
$\lambda_{Te}(c)$ and $\hat{\lambda}_{Te}(c)$ are as close as
possible\footnote{Consistently with most mathematical literature we
use the caret symbol (\^\/\/) to indicate estimation.}. What ``as
close as possible'' exactly means will be formalized by an appropriate
evaluation measure (see \S \ref{sec:evaluationmeasures}).

The reasons why we focus on \emph{binary} quantification are two-fold:

\begin{itemize}

\item Many quantification problems are binary in nature. For instance,
  estimating the prevalence of positive and negative reviews in a
  dataset of reviews of a given product is such a task.  Another such
  task is estimating from blog posts the prevalence of support for
  either of two candidates in the second round of a two-round
  (``run-off'') election.
 
\item A multi-class multi-label problem (also known as an
  \emph{$n$-of-$m$ problem}, i.e., a problem where zero, one, or
  several among $m$ classes can be attributed to the same document)
  can be reduced to $m$ independent binary problems of type ($c_{j}$
  vs.\ $\overline{c}_{j}$), where $\mathcal{C}=\{c_{1}, ..., c_{j},
  ..., c_{m}\}$ is the set of classes and where $\overline{c}_{j}$
  denotes the complement of $c_{j}$. Binary quantification methods can
  thus also be applied to solving quantification in multi-class
  multi-label contexts.

\end{itemize}

\noindent We instead leave the discussion of quantification in
single-label multi-class (i.e., 1-of-$m$) contexts to future work.


\subsection{Evaluation measures for quantification}
\label{sec:evaluationmeasures}

\noindent Different measures have been used in the literature for
measuring binary quantification accuracy.

The simplest such measure is \emph{bias} (B), defined as
$B(\lambda_{Te},\hat\lambda_{Te})=\hat{\lambda}_{Te}(c)-\lambda_{Te}(c)$
and used in \cite{Forman:2005fk,Forman06,Tang:2010uq}; positive bias
indicates a tendency to overestimate the prevalence of $c$, while
negative bias indicates a tendency to underestimate it.

\emph{Absolute Error} (AE - also used in \cite{Esuli:2010fk}, where it
is called \emph{percentage discrepancy}, and in
\cite{Barranquero:2013fk,Bella:2010kx,Forman:2005fk,Forman06,Gonzalez-Castro:2013fk,Sanchez:2008uq,Tang:2010uq}),
defined as
$AE(\lambda_{Te},\hat\lambda_{Te})=|\hat{\lambda}_{Te}(c)-\lambda_{Te}(c)|$,
is an alternative, equally simplistic measure that accounts for the
fact that positive and negative bias are (in the absence of specific
application-dependent constraints) equally undesirable.

\emph{Relative absolute error} (RAE), defined as
\begin{equation}\label{eq:RAE}
  RAE(\lambda_{Te},\hat\lambda_{Te})=\displaystyle\frac{|\hat{\lambda}_{Te}(c)-\lambda_{Te}(c)|}{\lambda_{Te}(c)}
\end{equation}
\noindent is a refinement of AE meant to account for the fact that the
same value of absolute error is a more serious mistake when the true
class prevalence is small. For instance, predicting
$\hat\lambda_{Te}(c)=0.10$ when $\lambda_{Te}(c)=0.01$ and predicting
$\hat\lambda_{Te}(c)=0.50$ when $\lambda_{Te}(c)=0.41$ are equivalent
errors according to $B$ and $AE$, but the former is intuitively a more
serious error than the latter.

The most convincing among the evaluation measures proposed so far is
certainly Forman's \citeyear{Forman:2005fk}, who uses \emph{normalized
cross-entropy}, better known as \emph{Kullback-Leibler Divergence}
(KLD -- see e.g., \cite{Cover91}). KLD, defined as
\begin{equation}\label{eq:KLD}
  KLD(\lambda_{Te},\hat\lambda_{Te})=\sum_{c\in C}\lambda_{Te}(c)\log \displaystyle\frac{\lambda_{Te}(c)}{\hat\lambda_{Te}(c)}
\end{equation}
\noindent and also used in
\cite{Esuli:2010fk,Forman06,Forman:2008kx,Tang:2010uq}, is a measure
of the error made in estimating a true distribution $\lambda_{Te}$
over a set $C$ of classes by means of a distribution
$\hat\lambda_{Te}$; this means that KLD is in principle suitable for
evaluating quantification, since quantifying exactly means predicting
how the test items are distributed across the classes. KLD ranges
between 0 (perfect coincidence of $\lambda_{Te}$ and
$\hat\lambda_{Te}$) and $+\infty$ (total divergence of $\lambda_{Te}$
and $\hat\lambda_{Te}$). In the binary case in which
$C=\{c,\overline{c}\}$, KLD becomes
\begin{equation}\label{eq:KLDbinary}KLD(\lambda_{Te},\hat\lambda_{Te})=\lambda_{Te}(c)\log\frac{\lambda_{Te}(c)}{\hat\lambda_{Te}(c)}
  +
  \lambda_{Te}(\overline{c})\log\frac{\lambda_{Te}(\overline{c})}{\hat\lambda_{Te}(\overline{c})}
\end{equation}
\noindent Continuity arguments indicate that we should consider $0
\log \frac{0}{q}=0$ and $p \log \frac{p}{0}=+\infty$ (see \cite[p.\
18]{Cover91}).  Note that, as from Equation \ref{eq:KLD}, KLD is
undefined when the predicted distribution $\hat\lambda_{Te}$ is zero
for at least one class (a problem that also affects RAE).  As a
result, we smooth the fractions
$\lambda_{Te}(c)/\hat{\lambda}_{Te}(c)$ and
$\lambda_{Te}(\overline{c})/\hat{\lambda}_{Te}(\overline{c})$ in
Equation \ref{eq:KLD} by adding a small quantity $\epsilon$ to both
the numerator and the denominator. The smoothed KLD function is always
defined and still returns a value of zero when $\lambda_{Te}$ and
$\hat\lambda_{Te}$ coincide.

KLD offers several advantages with respect to RAE (and, \textit{a
fortiori}, to B and AE). One advantage is that, as evident from
Equation \ref{eq:KLDbinary}, it is symmetric with respect to the
complement of a class, i.e., switching the role of $c$ and
$\overline{c}$ does not change the result. This means that, e.g.,
predicting $\hat\lambda_{Te}(c)=0.10$ when $\lambda_{Te}(c)=0.11$ and
predicting $\hat\lambda_{Te}(c)=0.90$ when $\lambda_{Te}(c)=0.89$, are
equivalent errors (which seems intuitive), while RAE considers the
former a much more serious error than the latter. This is especially
useful in binary quantification tasks in which it is not clear which
of the two classes should play the role of the positive class $c$, as
in e.g., \textsf{Employed} vs.\ \textsf{Unemployed}. A second
advantage is that KLD is not defined only on the binary (and
multi-label multi-class) case, but is also defined on the single-label
multi-class case; this allows evaluating different types of
quantification tasks with the same measure. Last but not least, one
benefit of using KLD is that it is a very well-known measure, having
been the subject of intense study within information theory
\cite{Csiszar:2004fk} and, although from a more applicative angle,
within the language modelling approach to information retrieval
\cite{Zhai:2008fk}.

\subsection{Existing quantification methods}
\label{sec:existingquantificationmethods}

\noindent A number of methods have been proposed in the (still brief)
literature on quantification; below we list the main ones, which we
will use as baselines in the experiments discussed in Section
\ref{sec:experiments}.

%
\textbf{Classify and Count (CC).} An obvious method for quantification
consists of generating a classifier from $Tr$, classifying the
documents in $Te$, and estimating $\lambda_{Te}$ by simply counting
the fraction of documents in $Te$ that are predicted positive, i.e.,
\begin{equation}\label{eq:classifyandcount}
  \hat{\lambda}_{Te}^{CC}(c)=\frac{|\{d_{j}\in Te|\hat{\Phi}(d_{j})=+1\}|}{|Te|}
\end{equation}
Forman \citeyear{Forman:2008kx} calls this the \emph{classify and
count} (CC) method.

\textbf{Probabilistic Classify and Count (PCC).} A variant of the
above consists in generating a classifier from $Tr$, classifying the
documents in $Te$, and computing $\lambda_{Te}$ as the \emph{expected}
fraction of documents predicted positive, i.e.,
\begin{equation}\label{eq:probclassifyandcount}
  \hat{\lambda}_{Te}^{PCC}(c)=\frac{1}{|Te|}\sum_{d_{j}\in
  Te}p(c|d_{j})
\end{equation}
\noindent where $p(c|d_{j})$ is the probability of membership in $c$
of test document $d_{j}$ returned by the classifier. If the classifier
only returns confidence scores that are not probabilities (as is the
case, e.g., when AdaBoost.MH is the learner \cite{Schapire00}), the
confidence scores must be converted into probabilities, e.g., by
applying a logistic function. The PCC method is dismissed as
unsuitable in \cite{Forman:2005fk,Forman:2008kx}, but is shown to
perform better than CC in \cite{Bella:2010kx} (where it is called
``Probability Average'') and in \cite{Tang:2010uq}.

%

\textbf{Adjusted Classify and Count (ACC).} Forman
\citeyear{Forman:2005fk,Forman:2008kx} uses a further method which he
calls ``Adjusted Count'', and which we will call \emph{Adjusted
Classify and Count} (ACC) so as to make its relation with CC more
explicit. The underlying idea is that CC would be optimal, were it not
for the fact that the classifier may generate different numbers of
false positives and false negatives, and that this difference would
lead to imperfect quantification. If we knew the ``true positive
rate'' ($tpr=\frac{TP}{TP+FN}$, a.k.a.\ recall) and ``false positive
rate'' ($fpr=\frac{FP}{FP+TN}$, a.k.a.\ fallout) that the classifier
has obtained on $Te$, it is easy to check that perfect quantification
would be obtained by adjusting $\hat{\lambda}_{Te}^{CC}(c)$ as
follows:
\begin{equation}\label{eq:ACC}
  \hat\lambda_{Te}^{ACC}(c)=\frac{\hat{\lambda}_{Te}^{CC}(c)-fpr_{Te}(c)}{tpr_{Te}(c)-fpr_{Te}(c)}
\end{equation}
\noindent Since we cannot know the true values of $tpr_{Te}(c)$ and
$fpr_{Te}(c)$, the ACC method consists of estimating them on $Tr$ via
$k$-fold cross-validation and using the resulting estimates in
Equation \ref{eq:ACC}.


However, one problem with ACC is that it is not guaranteed to return a
value in [0,1], due to the fact that the estimates of $tpr_{Te}(c)$
and $fpr_{Te}(c)$ may be imperfect. This lead Forman
\citeyear{Forman:2008kx} to ``clip'' the results of the estimation
(i.e., equate to 1 every value higher than 1 and to 0 every value
lower than 0) in order for the final results to be in [0,1].


\textbf{Probabilistic Adjusted Classify and Count (PACC).} The PACC
method (proposed in \cite{Bella:2010kx}, where it is called ``Scaled
Probability Average'') is a probabilistic variant of ACC, i.e., it
stands to ACC like PCC stands to CC. Its underlying idea is to
replace, in Equation \ref{eq:ACC}, $\hat{\lambda}_{Te}^{CC}(c)$,
$tpr_{Te}(c)$ and $fpr_{Te}(c)$ with their expected values, with
probability of membership in $c$ replacing binary
predictions. Equation \ref{eq:ACC} is thus transformed into
\begin{equation}\label{eq:PACC}
  \hat\lambda_{Te}^{PACC}(c)=\frac{\hat{\lambda}_{Te}^{PCC}(c)-E[fpr_{Te}(c)]}{E[tpr_{Te}(c)]-E[fpr_{Te}(c)]}
\end{equation}
\noindent where $E[tpr_{Te}(c)]$ and $E[fpr_{Te}(c)]$ (\emph{expected}
$tpr_{Te}(c)$ and \emph{expected} $fpr_{Te}(c)$, respectively) are
defined as
\begin{eqnarray}
  E[tpr_{Te}(c)] & = & \displaystyle\frac{1}{|Te_{c}|}\displaystyle\sum_{d_{j}\in Te_{c}}p(c|d_{j})\\
  E[fpr_{Te}(c)] & = &
  \displaystyle\frac{1}{|Te_{\overline{c}}|}\displaystyle\sum_{d_{j}\in
  Te_{\overline{c}}}p(c|d_{j})
\end{eqnarray}
\noindent and $Te_{c}$ (resp., $Te_{\overline{c}}$) indicates the set
of documents in $Te$ that belong (resp., do not belong) to class
$c$. Again, since we cannot know the true $E[tpr_{Te}(c)]$ and
$E[fpr_{Te}(c)]$ (given that we do not know $Te_{c}$ and
$Te_{\overline{c}}$), we estimate them on $Tr$ via $k$-fold
cross-validation and use the resulting estimates in Equation
\ref{eq:PACC}.
 

\textbf{Threshold$@$0.50 (T50), Method X (X), and Method Max (MAX).}
Forman \citeyear{Forman:2008kx} points out that the ACC method is very
sensitive to the decision threshold of the classifier, which may yield
unreliable values of $\lambda_{Te}^{ACC}(c)$ (or lead to
$\lambda_{Te}^{ACC}(c)$ being undefined when $tpr_{Te}=fpr_{Te}$). In
order to reduce this sensitivity, \cite{Forman:2008kx} recommends to
heuristically set the decision threshold in such a way that $tpr_{Tr}$
(as obtained via $k$-fold cross-validation) is equal to $.50$ before
computing Equation \ref{eq:ACC}. This method is dubbed
\emph{Threshold$@$0.50} (T50). Alternative heuristics that
\cite{Forman:2008kx} discusses are to set the decision threshold in
such a way that $fpr_{Tr}=1-tpr_{Tr}$ (this is dubbed \emph{Method X})
or such that $(tpr_{Tr}-fpr_{Tr})$ is maximized (this is dubbed
\emph{Method Max}).

\textbf{Median Sweep (MS).} Alternatively, \cite{Forman:2008kx}
recommends to compute $\lambda_{Te}^{ACC}(c)$ for every decision
threshold that gives rise (in $k$-fold cross-validation) to different
$tpr_{Tr}$ or $fpr_{Tr}$ values, and take the median of all the
resulting estimates of $\lambda_{Te}^{ACC}(c)$. This method is dubbed
\emph{Median Sweep} (MS).

\textbf{Mixture Model (MM).} The MM method (proposed in
\cite{Forman:2005fk}) consists of assuming that the distribution
$D^{Te}$ of the scores that the classifier assigns to the test
examples is a mixture
\begin{equation}\label{eq:MM}
  D^{Te} = \lambda_{Te}(c)\cdot D_{c}^{Te} + (1-\lambda_{Te}(c))\cdot D_{\overline{c}}^{Te}
\end{equation}
\noindent where $D_{c}^{Te}$ and $D_{\overline{c}}^{Te}$ are the
distributions of the scores that the classifier assigns to the
positive and the negative test examples, respectively, and where
$\lambda_{Te}(c)$ and $\lambda_{Te}(\overline{c})$ are the parameters
of this mixture. The MM method consists of estimating $D_{c}^{Te}$ and
$D_{\overline{c}}^{Te}$ via $k$-fold cross-validation, and picking as
value of $\lambda_{Te}(c)$ the one that generates the best fit between
the observed $D^{Te}$ and the mixture.  Two variants of this method,
called the \emph{Kolmogorov-Smirnov Mixture Model} (MM(KS)) and the
\emph{PP-Area Mixture Model} (MM(PP)), are actually defined in
\cite{Forman:2005fk}, which differ in terms of how the goodness of fit
between the left- and the right-hand side of Equation \ref{eq:MM} is
estimated. See \cite{Forman:2005fk} for more details.

\section{Optimizing quantification accuracy}
\label{sec:novelquantificationmethods}

\noindent A problem with the methods discussed in \S
\ref{sec:existingquantificationmethods} is that most of them are
fairly heuristic in nature. For instance, the fact that methods such
as ACC (and all the others based on it, such as T50, MS, X, and MAX)
require ``clipping'' is scarcely reassuring. More in general, methods
such as T50 or MS have hardly any theoretical foundation, and choosing
them over CC
only rests on our knowledge that they have performed better in
previously reported experiments.

A further problem is that some of these methods rest on assumptions
that seem problematic. For instance, one problem with the MM method is
that it seems to implicitly rely on the hypothesis that estimating
$D_{c}^{Te}$ and $D_{\overline{c}}^{Te}$ via $k$-fold cross-validation
on $Tr$ can be done reliably. However, since the very motivation of
doing quantification is that the training set and the test set may
have quite different characteristics, this hypothesis seems
adventurous. A similar argument casts some doubt on ACC: how reliable
are the estimates of $tpr_{Te}$ and $fpr_{Te}$ that can be generated
via $k$-fold cross-validation on $Tr$, given the different
characteristics that training set and test set may have in the
application contexts where quantification is required?\footnote{In
Appendix \ref{sec:ratioStudy} we thoroughly analyse (also by means of
concrete experiments) the issue of how (un)reliable the $k$-fold
cross-validation estimates of $tpr_{Te}$ and $fpr_{Te}$ are in
practice.} In sum, the very same arguments that are used to deem the
CC method unsuitable for quantification seem to undermine the
previously mentioned attempts at improving on CC.

In this paper we propose a new, theoretically well-founded
quantification method that radically differs from the ones discussed
in \S \ref{sec:existingquantificationmethods}.
Note that all of the methods discussed in \S
\ref{sec:existingquantificationmethods} employ \emph{general-purpose}
supervised learning methods, i.e., address quantification by
post-processing the results returned by a \emph{standard} classifier
(where the decision threshold has possibly been tuned according to
some heuristics). In particular, all the supervised learning methods
adopted in the literature on quantification optimize Hamming distance
or variants thereof, and not a quantification-specific evaluation
function. When the dataset is imbalanced (typically: when the
positives are by far outnumbered by the negatives), as is frequently
the case in text classification, this is suboptimal, since a
supervised learning method that minimizes Hamming distance will
generate classifiers with a tendency to make negative
predictions. This means that $FN$ will be much higher than $FP$, to
the detriment of quantification accuracy\footnote{\label{foot:ratio}To
witness, in the experiments we report in Section \ref{sec:experiments}
our 5148 test sets exhibit, when classified by the classifiers
generated by the linear SVM used for implementing the CC method, an
average $FP/FN$ ratio of 0.109; by contrast, for an optimal quantifier
this ratio is always 1.}.

We take a sharply different approach, based upon the use of
classifiers explicitly optimized for the evaluation function that we
will use for assessing quantification accuracy. Given such a
classifier, we will simply use a ``classify and count'' approach, with
no heuristic threshold tuning (\`a la T50 / X / MAX) and no \textit{a
posteriori} adjustment (\`a la ACC).


The idea of using learning algorithms capable of directly optimizing
the measure (a.k.a.\ ``loss'') used for evaluating effectiveness is
well-established in supervised learning. However, in our case
following this route is non-trivial, because the evaluation measure
that we want to use (KLD) is non-linear, i.e., is such that the error
on the test set may not be formulated as a linear combination of the
error incurred by each test example.  An evaluation measure for
quantification is \emph{inherently} non-linear,
because how the error on an individual test item impacts on the
overall quantification error depends on how the other test items have
been classified. For instance, if in the other test items there are
more false positives than false negatives, an additional false
negative is actually \emph{beneficial} to overall quantification
error, because of the mutual compensation effect between $FP$ and $FN$
mentioned in Section \ref{sec:introduction}. As a result, a measure of
quantification accuracy is inherently non-linear, and should thus be
multivariate, i.e., take in consideration all test items at once.

As discussed in \cite{Joachims05}, the assumption that the error on
the test set may be formulated as a linear combination of the error
incurred by each test example (as indeed happens for many common error
measures -- e.g., Hamming distance) underlies most existing
discriminative learners, which are thus suboptimal for tackling
quantification.  In order to sidestep this problem, we adopt the
\emph{SVM for Multivariate Performance Measures} (SVM$_{perf}$)
learning algorithm proposed by Joachims
\citeyear{Joachims05}\footnote{In \cite{Joachims05} SVM$_{perf}$ is
actually called $SVM{\Delta}_{multi}$, but the author has released its
implementation under the name SVM$_{perf}$. We will use this latter
name because it uniquely identifies the algorithm on the Web, while
searching for ``SVM multi'' often returns the $SVM^{multiclass}$
package, which addresses a different problem.}.  SVM$_{perf}$ is a
learning algorithm of the Support Vector Machine family that can
generate classifiers optimized for any non-linear, multivariate loss
function that can be computed from a contingency table (as KLD is).

SVM$_{perf}$ is a specialization to the problem of binary
classification of the \emph{structural SVM} ($SVM^{struct}$) learning
algorithm \cite{Joachims:2009kl,Joachims:2009fk,Tsochantaridis:2004jk}
for ``structured prediction'', i.e., an algorithm designed for
predicting multivariate, structured objects (e.g., trees, sequences,
sets).  SVM$_{perf}$ is fundamentally different from conventional
algorithms for learning classifiers: while these latter learn
univariate classifiers (i.e., functions of type
$\hat\Phi:D\rightarrow\{-1,+1\}$ that classify individual instances
one at a time), SVM$_{perf}$ learns \emph{multivariate} classifiers
(i.e., functions of type $\hat\Phi:D^{|S|}\rightarrow\{-1,+1\}^{|S|}$
that classify \emph{entire sets} $S$ of instances in one shot).
By doing so, SVM$_{perf}$ can optimize properties of entire sets of
instances, properties (such as KLD) that cannot be expressed as linear
functions of the properties of the individual instances.

As discussed in \cite{Joachims:2009fk}, $SVM^{struct}$ can be adapted
to a specific task by defining four components:

\begin{enumerate}
\item A \emph{joint feature map} $\Psi(\mathbf{x},\mathbf{y})$. This
  function computes a vector of features (describing the match between
  the input vectors in $\mathbf{x}$ and the relative outputs, true or
  predicted, in $\mathbf{y}$) from all the input-output pairs at the
  same time. In this way the number of features, and thus the number
  of parameters of the model, can be kept constant regardless of the
  size of the sample set. The $\Psi$ function allows to generalise not
  only on inputs ($\mathbf{x}$) but also on outputs ($\mathbf{y}$),
  thus allowing to produce predictions not seen in the training data.

  In SVM$_{perf}$ $\Psi$ is defined\footnote{For this formulation of
  $\Psi$, and when error rate
  is the chosen loss function, Joachims \citeyear{Joachims05} shows
  that SVM$_{perf}$ coincides with the traditional univariate SVM
  model (called SVM$_{org}$ in \cite{Joachims05}).}
  as
  \begin{equation}\label{eq:psiperf}
    \Psi(\mathbf{x},\mathbf{y})=\frac{1}{n}\sum_{i=1}^{n}y_i\mathbf{x_i}
  \end{equation}
\item A loss function $\Delta(\mathbf{y},\mathbf{\hat{y}})$.
  SVM$_{perf}$ works with loss functions $\Delta(TP,FP,FN,TN)$ in
  which the four values are those from the contingency table resulting
  from comparing the true labels $\mathbf{y}$ with the predicted
  labels $\mathbf{\hat{y}}$. In our work we take the loss function to
  be KLD, i.e.,\footnote{In Equation \ref{eq:lossKLD} $KLD$ is written
  as a function of $TP$, $FP$, $FN$, $TN$ for the simple fact that in
  \cite{Joachims05} (where SVM$_{perf}$ was originally described) the
  loss function $\Delta$ is specified as a function of the four cells
  of the contingency table.
  However, it should be clear that $\lambda(c)$ does not depend on the
  predicted labels: even if in Equation \ref{eq:lossKLD} we have
  written it out as $\lambda(c)=\frac{TP+FN}{TP+FP+FN+TN}$, this
  latter is equivalent to writing $\lambda(c)=\frac{GP}{GP+GN}$, where
  $GP$ (the ``gold positives'') is $TP+FN$ and $GN$ (the ``gold
  negatives'') is $FP+TN$. Seen under this light, there is no trace of
  predicted labels in $\lambda(c)=\frac{GP}{GP+GN}$, and $\lambda(c)$
  is just a function of the gold standard and not of the
  prediction. Analogously, it should be clear that $\hat\lambda(c)$ is
  just a function of the prediction and not of the gold standard.}
  \begin{equation}\label{eq:lossKLD}
    \Delta_{KLD}(TP,FP,FN,TN)=KLD(\lambda,\hat\lambda)
  \end{equation}
  \noindent where $\lambda(c)=\displaystyle\frac{TP+FN}{TP+FP+FN+TN}$
  and $\hat\lambda_{Te}(c)=\displaystyle\frac{TP+FP}{TP+FP+FN+TN}$

\item An algorithm for the efficient computation of a hypothesis
  \begin{equation}\label{eq:inferstruct}
    \hat\Phi(\mathbf{x})=\textrm{argmax}_{\mathbf{\hat{y}}\in \mathcal{Y}}\{\mathbf{w}\cdot\Psi(\mathbf{x},\mathbf{\hat{y}})\}
  \end{equation}
  \noindent where $\mathbf{w}$ is a vector of parameters. In
  SVM$_{perf}$ this simply corresponds to computing
  \begin{equation}\label{eq:inferperf}
    \hat\Phi(\mathbf{x})= (\textrm{sign}(\mathbf{w}\cdot x_1),\ldots,\textrm{sign}(\mathbf{w}\cdot x_n))
  \end{equation}
\item An algorithm for the efficient computation of the
  \emph{loss-augmented} hypothesis
  \begin{equation}\label{eq:inferlossstruct}
    \hat\Phi_{\Delta}(\mathbf{x})=\textrm{argmax}_{\mathbf{\hat{y}}\in \mathcal{Y}}\{\Delta(\mathbf{y},\mathbf{\hat{y}})+\mathbf{w}\cdot\Psi(\mathbf{x},\mathbf{\hat{y}})\}
  \end{equation}
  \noindent which in SVM$_{perf}$ is computed via an algorithm
  \cite[Algorithm 2]{Joachims05} with $O(n^2)$ worst-case complexity.
\end{enumerate}

\noindent We have used the implementation of SVM$_{perf}$ made
available by Joachims\footnote{SVM$_{perf}$is available from
\url{http://www.cs.cornell.edu/People/tj/svm\%5Flight/svm_perf.html}. Our
module that extends it to deal with KLD is available at
\url{http://hlt.isti.cnr.it/quantification/}.}, which we have extended
by implementing the module that takes care of the $\Delta_{KLD}$ loss
function. In the rest of the paper our method will be dubbed SVM(KLD).





\section{Experiments}
\label{sec:experiments}

\noindent We now present the results of experiments aimed at assessing
whether the approach we have proposed in Section
\ref{sec:novelquantificationmethods} delivers better quantification
accuracy than state-of-the-art quantification methods.  In order to do
this, we have run all our experiments by using as baselines for our
SVM(KLD) method all the methods described in \S
\ref{sec:existingquantificationmethods}. For the CC, ACC, T50, X, MAX,
MS, MM(KS), MM(PP) methods we have used the original implementation
that we have obtained from the author (this guarantees that the
baselines perform at their full potential). We have instead
implemented PCC and PACC ourselves.
At the heart of the implementation of all the baselines is a standard
linear SVM with the parameters set at their default values;
where quantities (such as e.g., $fpr_{Te}$ and $tpr_{Te}$ -- see
Equation \ref{eq:ACC}) had to be estimated from the training set, we
have used 50-fold cross-validation, as done and recommended in
\cite{Forman:2008kx}.  In order to guarantee a fair comparison with
the baselines we have used the default values for the parameters also
for SVM$_{perf}$, which lies at the basis of our SVM(KLD)
method\footnote{An additional reason why we have left the parameters
at their default values is that, in a context in which the
characteristics of $Tr$ and $Te$ may substantially differ, it is not
clear that the parameter values which are found optimal on $Tr$ via
$k$-fold cross-validation will also prove optimal (or at least will
perform reasonably) on $Te$.}.

In order to generate the vectorial representations for our documents,
the classic ``bag-of-words'' approach has been adopted. In particular,
punctuation has been removed, all letters have been converted to
lowercase, numbers have been removed, stop words have been removed
using the stop list provided in~\cite[pages 117--118]{Lewis92a}, and
stemming has been performed by means of the version of Porter's
stemmer available from \url{http://snowball.tartarus.org/}. All the
remaining stemmed words (``terms'') that occur at least once in $Tr$
have thus been used as features of our vectorial representations of
documents; no feature selection has been performed.  Feature weights
have been obtained via the ``ltc'' variant~\cite{Salton88} of the
well-known $tfidf$ class of weighting functions, i.e.,
\begin{equation} \label{eq:tfidf} tfidf(t_{k},d_{i}) =
  tf(t_{k},d_{i})\cdot \log \displaystyle\frac{|Tr|}{\#_{Tr}(t_{k})}
\end{equation}
\noindent where $d_{i}$ is a document, $\#_{Tr}(t_{k})$ denotes the
number of documents in $Tr$ in which feature $t_{k}$ occurs at least
once and
\begin{eqnarray}
  \begin{array}{lcl} tf(t_{k},d_{i}) & = & \left\{ \begin{array}{ll} 
        1+\log
        \#(t_{k},d_{i}) & \mbox{if } \#(t_{k},d_{i})>0 \\ 0 & \mbox{otherwise}
      \end{array} \right . \end{array}
\end{eqnarray}
\noindent where $\#(t_{k},d_{i})$ denotes the number of times $t_{k}$
occurs in $d_{i}$. Weights obtained by Equation~\ref{eq:tfidf} are
normalized through cosine normalization, i.e.,
\begin{eqnarray}\label{eq:normalizedtfidf} w_{ki}& = &
  \frac{tfidf(t_{k},d_{i})}{\sqrt{\sum_{s=1}^{|\mathcal{T}|}
  tfidf(t_{s},d_{i})^2}}\end{eqnarray}
%
\noindent where $\mathcal{T}$ denotes the total number of
features. Following \cite{Forman:2008kx}, we set the $\epsilon$
constant for smoothing KLD to the value $\epsilon=\frac{1}{2}|Te|$.

\subsection{Datasets}
\label{sec:datasets}

\noindent The datasets we use for our experiments have been extracted
from two important text classification test collections,
\textsc{Reuters Corpus Volume 1} version 2 (\textsc{RCV1-v2}) and
\textsc{OHSUMED-S}.

\textsc{RCV1-v2} is a standard, publicly available benchmark for text
classification consisting of 804,414 news stories produced by Reuters
from 20 Aug 1996 to 19 Aug
1997\footnote{\texttt{http://trec.nist.gov/data/reuters/reuters.html}}. \textsc{RCV1-v2}
ranks as one of the largest corpora currently used in text
classification research and, as pointed out in \cite{Forman:2006oa},
suffers from extensive ``drift'', i.e., from substantial variability
between the training set and the test set, which makes it a
challenging dataset for quantification. In our experiments we have
used the 12,807 news stories of the 1st week (20 to 26 Aug 1996) for
training, and the 791,607 news stories of the other 52 weeks for
testing\footnote{This is the standard ``LYRL2004'' split between
training and test data, originally defined in
\cite{Lewis:2004fk}.}. We have further partitioned these latter into
52 test sets each consisting of one week's worth of data\footnote{More
precisely, since the period covered by \textsc{RCV1-v2} consists of
365 days, i.e., 52 full weeks + 1 day, the 52nd test set consists of 1
day's worth of data only.}. \textsc{RCV1-v2} is multi-label, i.e., a
document may belong to several classes at the same time. Of the 103
classes of which its ``Topic'' hierarchy consists, in our experiments
we have restricted our attention to the 99 classes with at least one
positive training example. Consistently with the evaluation presented
in \cite{Lewis:2004fk}, also classes placed at internal nodes in the
hierarchically organized classification scheme are considered in the
evaluation; as positive examples of these classes we use the union of
the positive examples of their subordinate nodes, plus their ``own''
positive examples.

The \textsc{OHSUMED-S} dataset \cite{Esuli:2013ko} is a subset of the
well-known \textsc{OHSUMED} test collection \cite{Hersh94}.
\textsc{OHSUMED-S} consists of a set of 15,643 MEDLINE records
spanning the years from 1987 to 1991, where each record is classified
under one or more of the 97 MeSH index terms that belong to the
\emph{Heart Disease} (HD) subtree of the well-known MeSH tree of index
terms\footnote{\url{https://www.nlm.nih.gov/mesh/}}.  Each entry
consists of summary information relative to a paper published on one
of 270 medical journals; the available fields are title, abstract,
author, source, publication type, and MeSH index terms. As the
training set we have used, consistently with \cite{Esuli:2013ko}, the
2,510 documents belonging to year 1987; 9 MeSH index terms out of the
97 in the HD subtree are never assigned to any training document, so
the number of classes we actually use is 88.  We partition the four
remaining years' worth of data into four bins (1988, 1989, 1990,
1991), each containing the documents generated within the
corresponding calendar year. The reason why we do not use the entire
\textsc{OHSUMED} dataset is that roughly 93\% of \textsc{OHSUMED}
entries have no class assigned from the HD subtree, which means that
the classes in the HD subtree have very low prevalence
($\lambda_{Tr}=0.003$ on average); we thus prefer to use
\textsc{OHSUMED-S}, which presents a wider range of prevalence values.

This experimental setting thus generates $52\times99=5148$ binary
quantification test sets for \textsc{RCV1-v2} (containing an average
of 15,223 documents each), and $4\times88=352$ test sets for
\textsc{OHSUMED-S} (containing an average of 3,283 documents
each). This large number of test sets will give us an opportunity to
study quantification across different dimensions (e.g., across classes
characterized by different prevalence, across classes characterized by
different amounts of drift, across time)\footnote{In order to
guarantee perfect reproducibility of our results, we make available at
\url{http://hlt.isti.cnr.it/quantification/} the feature vectors of
the \textsc{RCV1-v2} and \textsc{OHSUMED-S} documents as extracted
from our preprocessing module, and already split respectively into the
53 and 5 sets described above.}.
More detailed figures about our datasets are given in Table
\ref{tab:datasets}. Note that both \textsc{RCV1-v2} and
\textsc{OHSUMED-S} classes are characterized by severe imbalance, as
can be noticed by the two ``Avg prevalence of the positive class''
rows of Table \ref{tab:datasets}, where both values are very far away
from the value of 0.5, which represents perfect balance. On a side
note, it is well-known that the ``bag-of-words'' extraction process
outlined a few paragraphs above gives rise to very high-dimensional
(albeit sparse) vectors; our case is no exception, and the
dimensionality of our vectors is 53,204 (\textsc{RCV1-v2}) and 11,286
(\textsc{OHSUMED-S}), respectively.

\begin{table}[t]
  \begin{center}
    \medskip
  \tbl{\label{tab:datasets}Main characteristics of the datasets used in our experiments.}{
    \begin{tabular}{|c||c||c|c|c|c|}
      \hline
      & & \hspace{1em}\textsc{RCV1-v2}\hspace{1em} & \textsc{OHSUMED-S} \\
      \hline
      \multirow{3}{*}{\begin{sideways}{\textsc{All}}\end{sideways}} & Total \# of docs & 804,414 & 15,643 \\
      & \# of classes (i.e., binary tasks) & 99 & 88 \\
      & Time unit used for split & week & year \\
      \hline
      \multirow{8}{*}{\begin{sideways}{\textsc{Training}}\end{sideways}} 
      & \# of docs & 12,807 & 2,510 \\
      & \# of features & 53,204 & 11,286 \\
      & Min \# of positive docs per class & 2 & 1 \\
      & Max \# of positive docs per class & 5,581 & 782 \\
      & Avg \# of positive docs per class & 397 & 55 \\
      & Min prevalence of the positive class & 0.0001 & 0.0004 \\
      & Max prevalence of the positive class & 0.4375 & 0.3116 \\
      & Avg prevalence of the positive class & 0.0315 & 0.0218 \\
      \hline
      \multirow{9}{*}{\begin{sideways}{\textsc{Test}}\end{sideways}} 
      & \# of docs & 791,607 & 13,133 \\
      & \# of test sets per class & 52 & 4 \\
      & Avg \# of test docs per set & 15,212 & 3,283 \\
      & Min \# of positive docs per class & 0 & 0 \\
      & Max \# of positive docs per class & 9,775 & 1,250 \\
      & Avg \# of positive docs per class & 494 & 69 \\
      & Min prevalence of the positive class & 0.0000 & 0.0000 \\
      & Max prevalence of the positive class & 0.5344 & 0.3532 \\
      & Avg prevalence of the positive class & 0.0323 & 0.0209 \\
      \hline
    \end{tabular}
    }
  \end{center}
\end{table}

Note that the experimental protocol we adopt is different from the one
adopted by Forman. In \cite{Forman:2005fk,Forman06,Forman:2008kx} he
proposes a protocol in which, given a training set $Tr$ and a test set
$Te$, several controlled experiments are run by artificially altering
class prevalences (i.e., by randomly removing predefined percentages
of the positives or of the negatives) either on $Tr$ or on $Te$. This
protocol is meant to test the robustness of the methods with respect
to different ``distribution drifts'' (i.e., differences between
$\lambda_{Tr}(c)$ and $\lambda_{Te}(c)$ of different magnitude) and
different class prevalence values. We prefer to opt for a different
protocol, one in which ``natural'' training and test sets are used,
without artificial alterations. The reason is that artificial
alterations may generate class prevalence values and/or distribution
drifts that are simply not realistic (e.g., a situation in which
$\lambda_{Tr}(c)=.40$ and $\lambda_{Te}(c)=.01$); conversely, focusing
on naturally occurring datasets forces us to come to terms with
realistic levels of class prevalence and/or distribution drift. We
will thus adopt the latter protocol in all the experiments discussed
in this paper, ``compensating'' for the absence of artificial
alterations by also studying (see \S
\ref{sec:resultsdivergencedimension}) the behaviour of our methods
separately on test sets characterized by different (and natural)
levels of distribution drift.


\subsection{Testing quantification accuracy}
\label{sec:results}

\noindent We have run our experiments by learning quantifiers for each
class $c$ on the respective training set and testing the quantifiers
separately on each of the test sets, using KLD as the evaluation
measure. We have done this for all the 99 classes $\times$ 52 weeks in
\textsc{RCV1-v2} and for all the 88 classes $\times$ 4 years in
\textsc{OHSUMED-S}, and for all the 10 baseline methods discussed in
\S \ref{sec:existingquantificationmethods} plus our SVM(KLD) method.


\subsubsection{Analysing the results along the class dimension}
\label{sec:resultsclassdimension}

\noindent We first discuss the results according to the class
dimension, i.e., by averaging the results for each \textsc{RCV1-v2}
class across the 52 test weeks and for each \textsc{OHSUMED-S} class
across the 4 years\footnote{Wherever in this paper we speak of
averaging accuracy results across different test sets, what we mean is
actually \emph{macroaveraging}, i.e., taking the accuracy results on
the individual test sets and computing their arithmetic mean. This is
sharply different from \emph{microaveraging}, i.e., merging the test
sets and computing a single accuracy figure on the merged set. Quite
obviously, in a quantification setting microaveraging does not make
any sense at all, since false positives from one set and false
negatives from another set would compensate each other, thus
generating misleadingly high accuracy values.}. Since this would leave
no less than 99 \textsc{RCV1-v2} classes to discuss, we further
average the results across all the \textsc{RCV1-v2} classes
characterized by a training class prevalence $\lambda_{Tr}(c)$ that
falls into a certain interval (same for \textsc{OHSUMED-S} and its 88
classes). This allows us to separately check the behaviour of our
quantification methods on groups of classes that are homogeneous by
level of imbalance. We have also run statistical significance tests in
order to check whether the improvement obtained by the best performing
method over the 2nd best performer on the group is statistically
significant\footnote{All the statistical significance tests discussed
in this paper are based on a two-tailed paired t-test and the use of a
0.001 significance level.}.

The results are reported in Table \ref{tab:quantificationaccuracy},
where four levels of imbalance have been singled out: very low
prevalence (VLP, which accounts for all the classes $c$ such that
$\lambda_{Tr}(c)<0.01$; there are 48 \textsc{RCV1-v2} and 51
\textsc{OHSUMED-S} such classes), low prevalence (LP -- $0.01\leq
\lambda_{Tr}(c)<0.05$; 34 \textsc{RCV1-v2} and 28 \textsc{OHSUMED-S}
classes), high prevalence (HP -- $0.05\leq \lambda_{Tr}(c)<0.10$; 10
\textsc{RCV1-v2} and 4 \textsc{OHSUMED-S} classes), and very high
prevalence (VHP -- $0.10\leq \lambda_{Tr}(c)$; 7 \textsc{RCV1-v2} and
5 \textsc{OHSUMED-S} classes).

\begin{table}[t]
  \tbl{Accuracy of SVM(KLD) and of 10 baseline methods as measured in terms of KLD on the 99 classes of \textsc{RCV1-v2} (top) and on the 88 classes of \textsc{OHSUMED-S} (bottom) grouped by class prevalence in $Tr$ (Columns 2 to 5); lower values are better; Column 6 indicates average accuracy across all the classes. The best result in each column is indicated with \textbf{boldface} only when there is a statistically significant difference with respect to each of the other tested methods ($p<0.001$, two-tailed paired t-test on KLD value across the test sets in the group). The methods are ranked in terms of the value indicated in the ``All'' column.}{
    \begin{tabular}{|c|c||c|c|c|c||c|}
      \hline
      &  & VLP & LP & HP & VHP & All \\
      \hline
      \multirow{11}{*}{\begin{sideways}\textsc{RCV1-v2}\end{sideways}}
      & SVM(KLD) & 2.09E-03 & \bf 4.92E-04 & 7.19E-04 & 1.12E-03 & \bf 1.32E-03 \\
      & PACC &	2.16E-03 &	1.70E-03 &	4.24E-04 &	2.75E-04 &	1.74E-03 \\
      & ACC & 2.17E-03 & 1.98E-03 & 5.08E-04 & 6.79E-04 & 1.87E-03 \\
      & MAX & 2.16E-03 & 2.48E-03 & 6.70E-04 & \bf 9.03E-05 & 2.03E-03 \\
      & CC & 2.55E-03 & 3.39E-03 & 1.29E-03 & 1.61E-03 & 2.71E-03 \\
      & X & 3.48E-03 & 8.45E-03 & 1.32E-03 & 2.43E-04 & 4.96E-03 \\
      & PCC & 1.04E-02 &	6.49E-03 &	3.87E-03 &	1.51E-03 &	7.86E-03 \\
      & MM(PP) & 1.76E-02 & 9.74E-03 & 2.73E-03 & 1.33E-03 & 1.24E-02 \\
      & MS & 1.98E-02 & 7.33E-03 & 3.70E-03 & 2.38E-03 & 1.27E-02 \\
      & T50 & 1.35E-02 & 1.74E-02 & 7.20E-03 & 3.17E-03 & 1.38E-02 \\
      & MM(KS) & 2.00E-02 & 1.14E-02 & 9.56E-04 & 3.62E-04 & 1.40E-02 \\
      \hline
      \hline
      &  & VLP & LP & HP & VHP & All \\
      \hline
      \multirow{11}{*}{\begin{sideways}\textsc{OHSUMED-S}\end{sideways}}
      & SVM(KLD)	&	\bf 1.21E-03	&	\bf 1.02E-03	&	5.55E-04	&	1.05E-03	& \bf	1.13E-03	\\
      & PACC	&	2.86E-03	&	2.78E-03	&	2.82E-04	&	4.76E-04	&	2.61E-03	\\
      & ACC	&	2.37E-03	&	5.40E-03	&	2.82E-04	&	\bf 2.57E-04	&	2.99E-03	\\
      & CC	&	2.38E-03	&	8.89E-03	&	2.82E-03	&	2.20E-03	&	4.12E-03	\\
      & X	&	1.38E-03	&	3.94E-03	&	3.35E-04	&	5.36E-03	&	4.44E-03	\\
      & MM(PP)	&	4.90E-03	&	1.41E-02	&	9.72E-04	&	4.94E-03	&	7.63E-03	\\
      & MM(KS)	&	1.37E-02	&	2.32E-02	&	8.42E-04	&	5.73E-03	&	1.14E-02	\\
      & MS	&	3.80E-03	&	1.79E-03	&	1.45E-03	&	1.90E-02	&	1.18E-02	\\
      & T50	&	7.53E-02	&	5.17E-02	&	2.71E-03	&	1.27E-02	&	2.74E-02	\\
      & MAX	&	5.57E-03	&	2.33E-02	&	1.76E-01	&	3.78E-01	&	3.67E-02	\\
      & PCC	&	1.20E-01	&	8.98E-02	&	4.93E-02	&	2.27E-02	&	1.04E-01	\\
      \hline
    \end{tabular}
}
  \label{tab:quantificationaccuracy}
\end{table}

The first observation that can be made by looking at the
\textsc{RCV1-v2} results in this table is that, when evaluated across
all our 5148 test sets (Column 6), SVM(KLD) outperforms all the other
baseline methods in a statistically significant way, scoring a KLD
value of 1.32E-03 against the 1.74E-03 value (a -24.2\% error
reduction) obtained by the best-performing baseline (the PACC
method). This is largely a result of a much better balance between
false positives and false negatives obtained by the base classifiers:
while (as already observed in Footnote \ref{foot:ratio}) the average
$FP/FN$ ratio across the 5148 test sets is 0.109 for CC, it is 0.684
for SVM(KLD) (and it is 1 for the perfect quantifier). The
\textsc{OHSUMED-S} results essentially confirm the insights obtained
from the \textsc{RCV1-v2} results, with SVM(KLD) again the best of the
11 methods and PACC again the 2nd best; the difference between them is
now even higher, with an error reduction of -56.8\%.

A second observation that the \textsc{RCV1-v2} results allow is to
make is that SVM(KLD) scores well on all the four groups of classes
identified; while it is not always the best method (e.g., it is
outperformed by other methods in the HP and VHP groups), it
consistently performs well on all four groups.  In particular,
SVM(KLD) seems to excel at classes characterized by drastic imbalance,
as witnessed by the VLP group, where SVM(KLD) is the best performer,
and by the LP group, where SVM(KLD) is the best performer in a
statistically significant way. In fact, this group of classes seems
largely responsible for the excellent overall performance (Column 6)
displayed by SVM(KLD), since on the LP group the margin between it and
the other methods is large (4.92E-04 against 1.98E-03 of the 2nd best
method), and since the VLP and LP groups altogether account for no
less than 82 out of the total 99 classes. This latter fact
characterizes most naturally occurring datasets, whose class
distribution usually exhibits a power law, with very few highly
frequent classes and very many highly infrequent classes. The
\textsc{OHSUMED-S} results essentially confirm the \textsc{RCV1-v2}
results, with SVM(KLD) again the best performer on the VLP and LP
classes, and still performing well, although not being the best, on HP
and VHP.

The stability of SVM(KLD) is also confirmed by Table
\ref{tab:quantificationvariance}, which reports, for the same groups
of test sets identified by Table \ref{tab:quantificationaccuracy}, the
variance in KLD across the members of the group. For example, on
\textsc{RCV1-v2}, Column 3 reports the variance in KLD across all the
34 classes such that $.01 \leq \lambda_{Tr}(c) \leq 0.05$ and across
the 52 test weeks, for a total of $34\times52=1,768$ test sets.  What
we can observe from this table is that, when averaged across all the
$99\times52=5148$ test sets (Column 6), the variance of SVM(KLD) is
lower than the variance of all other methods in a statistically
significant way. The variance of SVM(KLD) is fairly low in all the
four subsets of classes, and particularly so in the subsets of the
most imbalanced classes (VLP and LP), in which SVM(KLD) is the best
performer in a statistically significant way. The \textsc{OHSUMED-S}
results essentially confirm the \textsc{RCV1-v2} results, with
SVM(KLD) the best performer on VLP and LP, and still behaving well on
HP and VHP.

Concerning the baselines, our results seem to disconfirm the ones
reported in \cite{Forman:2008kx} according to which the MS method is
the best of the lot, and according to which the ACC method ``can
estimate the class distribution well in many situations, but its
performance degrades severely when the training class distribution is
highly imbalanced''.  In our experiments, instead, MS is substantially
outperformed by several baseline methods; ACC is instead a strong
contender, and (contrary to the statement above) especially shines on
the subsets of the most imbalanced classes. The results of both Tables
\ref{tab:quantificationaccuracy} and \ref{tab:quantificationvariance}
clearly show that, on both \textsc{RCV1-v2} and \textsc{OHSUMED-S},
PACC is the best of the baseline methods presented in \S
\ref{sec:existingquantificationmethods}.

\begin{table}[t]
  \tbl{Variance of SVM(KLD) and of  10 baseline methods as measured in terms of KLD on the 99 classes of \textsc{RCV1-v2} (top) and on the 88 classes of \textsc{OHSUMED-S} (bottom) grouped by class prevalence in $Tr$ (Columns 2 to 5); Column 6 indicates variance across all the classes. \textbf{Boldface} represents the best value. The methods are ranked in terms of the value indicated in the ``All'' column.}{
    \begin{tabular}{|c|c||c|c|c|c||c|}
      \hline
      & & VLP & LP & HP & VHP & All \\
      \hline
      \multirow{11}{*}{\begin{sideways}\textsc{RCV1-v2}\end{sideways}}
      & SVM(KLD)	& \bf 7.52E-06 & \bf 3.44E-06 & 8.94E-07 & 1.56E-06 & \bf 5.68E-06 \\
      & PACC &	7.58E-06 &	2.38E-05 &	1.50E-06 &	2.26E-07 &	1.29E-05 \\
      & ACC & 1.04E-05 & 7.43E-06 & \bf 4.25E-07 & 4.26E-07 & 8.18E-06 \\
      & MAX & 8.61E-06 & 2.27E-05 & 1.06E-06 & \bf 1.66E-08 & 1.32E-05 \\
      & CC & 1.79E-05 & 1.99E-05 & 1.96E-06 & 1.66E-06 & 1.68E-05 \\
      & X & 2.21E-05 & 6.57E-04 & 2.28E-06 & 1.06E-07 & 2.64E-04 \\
      & PCC &	1.75E-04 &	1.76E-04 &	3.56E-05 &	1.59E-04 & 	9.38E-04\\
      & T50 & 2.65E-04 & 4.56E-04 & 2.43E-04 & 1.19E-05 & 3.33E-04 \\
      & MM(KS) & 3.65E-03 & 7.81E-04 & 1.46E-06 & 4.43E-07 & 2.10E-03 \\
      & MM(PP) & 4.07E-03 & 5.69E-04 & 6.35E-06 & 2.66E-06 & 2.21E-03 \\
      & MS & 9.36E-03 & 5.80E-05 & 1.31E-05 & 6.18E-06 & 4.61E-03 \\
      \hline
      \hline
      & & VLP & LP & HP & VHP & All \\
      \hline
      \multirow{11}{*}{\begin{sideways}\textsc{OHSUMED-S}\end{sideways}}
      & SVM(KLD)	&	\bf 3.33E-06	&	\bf 8.95E-06	&	3.74E-07	&	3.73E-06	&	\bf 4.73E-06	\\
      & PACC	&	9.01E-05	&	4.88E-05	&	9.73E-08	&	4.85E-07	&	7.12E-05	\\
      & ACC	&	1.06E-05	&	1.19E-04	&	\bf 8.92E-08	&	\bf 1.34E-07	&	4.08E-05	\\
      & CC	&	1.06E-05	&	1.15E-04	&	2.74E-06	&	3.41E-06	&	4.58E-05	\\
      & X	&	4.72E-06	&	1.45E-04	&	1.15E-07	&	8.72E-05	&	9.85E-05	\\
      & MM(PP)	&	5.79E-05	&	5.65E-04	&	1.93E-06	&	8.38E-05	&	2.50E-04	\\
      & MM(KS)	&	2.24E-03	&	1.02E-03	&	7.38E-07	&	1.18E-04	&	5.55E-04	\\
      & MS	&	2.86E-05	&	6.43E-06	&	4.23E-06	&	2.20E-03	&	1.35E-03	\\
      & T50	&	1.77E-02	&	2.83E-03	&	1.78E-05	&	2.21E-04	&	2.23E-03	\\
      & MAX	&	4.38E-04	&	5.22E-03	&	5.28E-02	&	2.89E-01	&	2.60E-02	\\
      & PCC	&	6.78E-05	&	9.64E-05	&	3.78E-05	&	2.36E-04	&	7.63E-04	\\
      \hline
    \end{tabular}
     }
  \label{tab:quantificationvariance}
\end{table}



\subsubsection{Analysing the results along the temporal dimension}
\label{sec:resultstemporaldimension}

\begin{figure}[t]
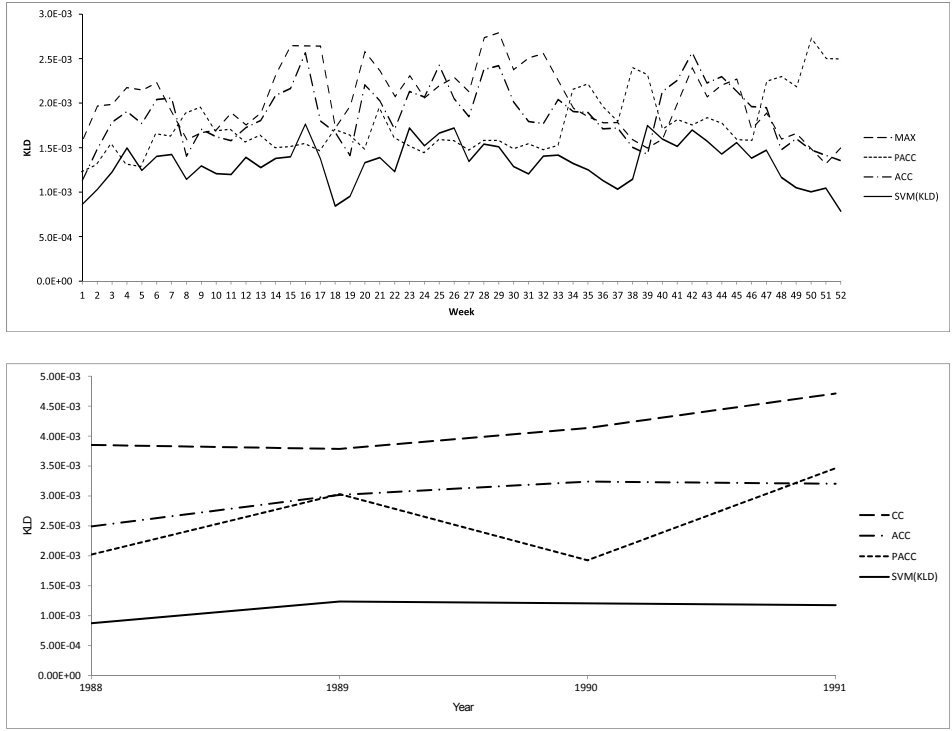

  \caption{\label{fig:KLDTimePlot}Values of KLD obtained by SVM(KLD)
  and by the three best-performing baseline methods on the
  \textsc{RCV1-v2} dataset (top) and on the \textsc{OHSUMED-S} dataset
  (bottom). Each point represents the average KLD value obtained
  across the 99 \textsc{RCV1-v2} classes for the given week (top) and
  across the 88 \textsc{OHSUMED-S} classes for the given year
  (bottom); lower values are better. Points are plotted as a temporal
  series (the 52 weeks and the 4 years, respectively, are
  chronologically ordered). \\ \hspace{1em}}
  \begin{center}
    \scalebox{.48}[.48]{\includegraphics{KLDTimePlot}} \\ \mbox{} \\
    \scalebox{.48}[.48]{\includegraphics{KLDTimePlotOHS}} \\
  \end{center}
\end{figure}

\noindent We now analyse the results along the temporal dimension. In
order to do this for the \textsc{RCV1-v2} dataset (resp.,
\textsc{OHSUMED-S} dataset), for each of the 52 test weeks (resp., 4
test years) we average the 99 (resp., 88) accuracy results
corresponding to the individual classes, and check the temporal
accuracy trend resulting from these averages. This trend is displayed
in Figure \ref{fig:KLDTimePlot}, where the results of SVM(KLD) are
plotted together with the results of the three best-performing
baseline methods. The plots unequivocally show that SVM(KLD) is the
best method across the entire temporal spectrum for both
\textsc{RCV1-v2} and \textsc{OHSUMED-S}.

Note that quantification accuracy remains fairly stable across time,
i.e., we are not witnessing any substantial decrease in quantification
accuracy with time. Intuition might instead suggest that
quantification accuracy should decrease with time, due to the combined
effects of true concept drift and distribution drift. This may
indicate that (at least in the context of broadcast news that
\textsc{RCV1-v2} represents, and in the context of medical scientific
articles that \textsc{OHSUMED-S} represents) the chosen timeframe (one
year for \textsc{RCV1-v2}, four years for {OHSUMED-S}) is not
sufficient enough a timeframe to observe a significant such drift.

\subsubsection{Analysing the results along the distribution drift
dimension}
\label{sec:resultsdivergencedimension}

\noindent The last angle according to which we analyse the results is
distribution ``drift''. That is, we conduct our analysis in terms of
how much the prevalence $\lambda_{Te_{i}}(c)$ in a given test set
$Te_{i}$ ``drifts away'' from the prevalence $\lambda_{Tr}(c)$ in the
training set. Specifically, for all our 5148 \textsc{RCV1-v2} test
sets (resp., 352 \textsc{OHSUMED-S} test sets) $Te_{i}$ we compute
$KLD(\lambda_{Te_{i}},\lambda_{Tr})$, we rank all the test sets
according to the KLD value they have obtained, and we subdivide the
resulting ranking into four equally-sized segments (quartiles) of
5148/4=1287 \textsc{RCV1-v2} test sets (resp., 352/4=88
\textsc{OHSUMED-S} test sets) each. As a result, each resulting
quartile contains test sets that are homogeneous according to the
divergence of their distributions from the corresponding distributions
in the training set (different test sets pertaining to the same class
$c$ may thus end up in different quartiles).
This allows us to investigate how well the different methods behave
when distribution drift is low (indicated by low KLD values) and when
distribution drift is high (high KLD values). We have also run
statistical significance tests in order to check whether the
improvement obtained by the best performing method over the 2nd best
performer on the test sets belonging to a certain quartile is
statistically significant.

\begin{table}[t]
  \tbl{As Table \protect\ref{tab:quantificationaccuracy}, but with tests sets grouped by distribution drift instead of by class prevalence.}{
    \begin{tabular}{|c|c||c|c|c|c||c|}
      \hline
      & & VLD & LD & HD & VHD & All \\
      \hline
      \multirow{11}{*}{\begin{sideways}\textsc{RCV1-v2}\end{sideways}}
      & SVM(KLD) & \bf 1.17E-03 & \bf 1.10E-03 &  \bf 1.38E-03 & 1.67E-03 & \bf 1.32E-03 \\
      & PACC &	1.92E-03 &	2.11E-03 &	1.74E-03 &	1.20E-03 &	1.74E-03 \\
      & ACC & 1.70E-03 & 1.74E-03 & 1.93E-03 & 2.14E-03 & 1.87E-03 \\
      & MAX & 2.20E-03 & 2.15E-03 & 2.25E-03 & 1.52E-03 & 2.03E-03 \\
      & CC & 2.43E-03 & 2.44E-03 & 2.79E-03 & 3.18E-03 & 2.71E-03 \\
      & X & 3.89E-03 & 4.18E-03 & 4.31E-03 & 7.46E-03 & 4.96E-03 \\
      & PCC &	8.92E-03 &	8.64E-03 &	7.75E-03 &	6.24E-03 &	7.86E-03\\
      & MM(PP) & 1.26E-02 & 1.41E-02 & 1.32E-02 & 1.00E-02 & 1.24E-02 \\
      & MS & 1.37E-02 & 1.67E-02 & 1.20E-02 & 8.68E-03 & 1.27E-02 \\
      & T50 & 1.17E-02 & 1.38E-02 & 1.49E-02 & 1.50E-02 & 1.38E-02 \\
      & MM(KS) & 1.41E-02 & 1.58E-02 & 1.53E-02 & 1.10E-02 & 1.40E-02 \\
      \hline
      \hline
      & & VLD & LD & HD & VHD & All \\
      \hline
      \multirow{11}{*}{\begin{sideways}\textsc{OHSUMED-S}\end{sideways}}
      & SVM(KLD)	& \bf	7.00E-04	& \bf	7.54E-04	&  \bf	9.39E-04	&	2.11E-03	& \bf	1.13E-03	\\
      & PACC	&	2.91E-03	&	3.60E-03	&	3.01E-03	& \bf	9.25E-04	&	2.61E-03	\\
      & ACC	&	2.54E-03	&	2.83E-03	&	2.46E-03	&	4.14E-03	&	2.99E-03	\\
      & CC	&	3.31E-03	&	3.87E-03	&	3.87E-03	&	5.43E-03	&	4.12E-03	\\
      & X	&	5.59E-03	&	5.74E-03	&	3.83E-03	&	2.61E-03	&	4.44E-03	\\
      & MM(PP)	&	7.82E-03	&	7.21E-03	&	7.35E-03	&	8.14E-03	&	7.63E-03	\\
      & MM(KS)	&	1.10E-02	&	1.14E-02	&	8.66E-03	&	1.44E-02	&	1.14E-02	\\
      & MS	&	1.52E-02	&	9.88E-03	&	1.50E-02	&	7.11E-03	&	1.18E-02	\\
      & T50	&	2.25E-02	&	3.07E-02	&	2.30E-02	&	3.32E-02	&	2.74E-02	\\
      & MAX	&	5.00E-02	&	2.88E-02	&	2.69E-02	&	4.10E-02	&	3.67E-02	\\
      & PCC	&	1.10E-01	&	1.07E-01	&	9.97E-02	&	9.88E-02	&	1.04E-01	\\
      \hline
    \end{tabular}
     }
  \label{tab:resultsaccordingtodivergence}
\end{table}

\begin{table}[t]
  \tbl{As Table \protect\ref{tab:quantificationvariance}, but with tests sets grouped by distribution drift instead of by class prevalence.}{
    \begin{tabular}{|c|c||c|c|c|c||c|}
      \hline
      & & VLD & LD & HD & VHD & All \\
      \hline
      \multirow{11}{*}{\begin{sideways}\textsc{RCV1-v2}\end{sideways}}
      & SVM(KLD) & \bf 2.73E-06 & \bf 2.08E-06 & \bf 3.91E-06 & 1.38E-05 & \bf 5.68E-06 \\
      & PACC &	9.71E-06  &	2.26E-05  &	1.15E-05 & \bf	7.14E-06 &	1.29E-05 \\
      & ACC & 6.20E-06 & 6.67E-06 & 7.97E-06 & 1.18E-05 & 8.18E-06 \\
      & MAX & 1.45E-05 & 1.32E-05 & 1.57E-05 & 9.19E-06 & 1.32E-05 \\
      & CC & 1.57E-05 & 1.57E-05 & 1.70E-05 & 1.85E-05 & 1.68E-05 \\
      & X & 1.22E-04 & 1.30E-04 & 1.28E-04 & 6.71E-04 & 2.65E-04 \\
      & PCC &	7.61E-04 &	8.38E-04 &	8.24E-04 &	8.85E-04  &	9.38E-04 \\
      & T50 & 2.13E-04 & 2.42E-04 & 2.64E-04 & 6.08E-04 & 3.33E-04 \\
      & MM(KS) & 2.12E-03 & 2.11E-03 & 2.94E-03 & 1.25E-03 & 2.11E-03 \\
      & MM(PP) & 2.70E-03 & 3.06E-03 & 1.93E-03 & 1.18E-03 & 2.22E-03 \\
      & MS & 4.97E-03 & 9.03E-03 & 2.38E-03 & 2.05E-03 & 4.61E-03 \\      \hline
      \hline
      & & VLD & LD & HD & VHD & All \\
      \hline
      \multirow{11}{*}{\begin{sideways}\textsc{OHSUMED-S}\end{sideways}}
      & SVM(KLD)&	\bf 7.59E-07	&	\bf 8.31E-07	&	\bf 2.47E-06	&	1.37E-05	&	\bf 4.73E-06	\\
      & PACC	&	1.90E-05	&	1.59E-04	&	1.01E-04	&	\bf 4.16E-06	&	7.12E-05	\\
      & ACC 	&	1.49E-05	&	2.27E-05	&	3.35E-05	&	9.16E-05	&	4.08E-05	\\
      & CC  	&	1.23E-05	&	2.88E-05	&	2.60E-05	&	1.15E-04	&	4.58E-05	\\
      & X	&	9.03E-05	&	1.95E-04	&	6.46E-05	&	4.07E-05	&	9.85E-05	\\
      & MM(PP)	&	2.37E-04	&	2.50E-04	&	1.57E-04	&	3.65E-04	&	2.50E-04	\\
      & MM(KS)	&	4.24E-04	&	5.04E-04	&	1.91E-04	&	1.10E-03	&	5.55E-04	\\
      & MS	&	1.57E-03	&	1.20E-03	&	1.81E-03	&	8.01E-04	&	1.35E-03	\\
      & T50	&	1.05E-03	&	1.84E-03	&	1.26E-03	&	4.76E-03	&	2.23E-03	\\
      & MAX  	&	5.38E-02	&	7.31E-03	&	2.06E-02	&	2.27E-02	&	2.60E-02	\\
      & PCC 	&	5.86E-04	&	4.92E-04	&	9.21E-04	&	9.86E-04	&	7.63E-04	\\
      \hline   \end{tabular}
     }
  \label{tab:varianceaccordingtodivergence}
\end{table}

The results of this analysis are displayed in Table
\ref{tab:resultsaccordingtodivergence}. The most important observation
here is that SVM(KLD) is the best performer in a statistically
significant way, both overall and on three out of four quartiles (on
the ``very high drift'' quartile SVM(KLD) is outperformed by
PACC). Additionally, it is worth observing that its performance is
consistently high on each quartile. Table
\ref{tab:varianceaccordingtodivergence} reports variance figures,
showing again the stability of SVM(KLD), which is the most stable
method overall and on three out of four quartiles (on the VHD quartile
the most stable method is PACC).



\subsubsection{Evaluating the results according to RAE}
\label{sec:rae}

\noindent It may be interesting to analyse the results of the previous
experiments according to an evaluation measure different from KLD,
such as the RAE measure introduced in \S
\ref{sec:evaluationmeasures}. As discussed in \S
\ref{sec:evaluationmeasures}, RAE is the most (and only) reasonable
alternative to KLD proposed so far. Given the simplicity of its
mathematical form, it is also a measure everyone can relate to.

Similarly to Table \ref{tab:resultsaccordingtodivergence}, Table
\ref{tab:raevalues} reports the results of our experiments broken down
into quartiles of test sets homogeneous by distribution drift; the
difference with Table \ref{tab:resultsaccordingtodivergence} is that
RAE is now used as the evaluation measure in place of KLD. The
\textsc{RCV1-v2} results of Table
\ref{tab:resultsaccordingtodivergence} confirm the superiority of
SVM(KLD) over the baselines, notwithstanding the discrepancy between
evaluation measure (RAE) and loss function optimized (KLD). As an
average across the 5148 test sets, SVM(KLD) obtains an average RAE
value of 0.465, which improves in a statistically significant way over
the 0.674 result obtained by the second best performer, ACC; the next
best-performing methods, CC and PACC, obtain dramatically worse
results (1.087 and 1.466, respectively). SVM(KLD) is also the best
performer, in a statistically significant way, in each of the four
quartiles. 

In this case \textsc{OHSUMED-S} results are substantially different
from the \textsc{RCV1-v2} results, since here the best performer is
MM(KS) (a weak contender in all the experiments that we have reported
so far), while SVM(KLD) performs much less well. The overall
performance of SVM(KLD) is penalized by a bad performance on the VHD
quartile, since it is the best performer (and in a statistically
significant way) on the other three quartiles. While there is some
discrepancy between the outcomes of the \textsc{RCV1-v2} experiments
and those of the \textsc{OHSUMED-S} experiments, we observe that the
former might be considered somehow more trustworthy than the latter
ones, since \textsc{RCV1-v2} is much bigger than \textsc{OHSUMED-S}
(approximately 60 times more test documents).

Additionally, and more importantly, let us recall that SVM(KLD) is
optimized for KLD, and not for RAE. This means that, in keeping with
our plan to use classifiers explicitly optimized for the evaluation
function used for assessing quantification accuracy, should we really
deem RAE to be the ``right'' such function we would implement and use
SVM(RAE)!, and not SVM(KLD). In other words, in the \textsc{RCV1-v2}
results of Table \ref{tab:raevalues} SVM(KLD) turns out to be the best
performer \emph{despite the fact} that we here use RAE as an
evaluation measure.

\begin{table}[t]
  \tbl{As Table \protect\ref{tab:quantificationaccuracy}, but with RAE used as the evaluation measure instead of KLD.}{
    \begin{tabular}{|c|c||r|r|r|r||r|}
      \hline
      \multirow{11}{*}{\begin{sideways}\textsc{RCV1-v2}\end{sideways}}
      & & VLD \hspace{0.25em} & LD \hspace{0.25em} & HD \hspace{0.25em} & VHD \hspace{0.05em} & All \hspace{0.25em} \\
      \hline
      & SVM(KLD) & \bf 0.567 & \bf 0.504 & \bf 0.462 & \bf 0.328 & \bf 0.465 \\
      & ACC & 0.885 & 0.761 & 0.629 & 0.420 & 0.674 \\
      & CC & 1.433 & 1.174 & 1.059 & 0.683 & 1.087 \\
      & PACC &	2.235 &	1.924 &	1.369 &	0.402 &	1.466\\
      & MAX & 2.199 & 1.420 & 1.213 & 6.579 & 2.853 \\
      & X & 2.339 & 1.627 & 1.360 & 6.726 & 3.013 \\
      & MM(PP) & 6.506 & 6.189 & 5.258 & 3.378 & 5.333 \\
      & T50 & 5.804 & 5.238 & 4.667 & 6.674 & 5.596 \\
      & MM(KS) & 7.195 & 7.375 & 5.954 & 4.113 & 6.160 \\
      & PCC &	13.989 &	 16.258 &	5.788 &	2.017 &	9.433 \\
      & MS & 48.139 & 19.114 & 9.770 & 136.755 & 53.444 \\      \hline
      \hline
      \multirow{11}{*}{\begin{sideways}\textsc{OHSUMED-S}\end{sideways}}
      & & VLD \hspace{0.25em} & LD \hspace{0.25em} & HD \hspace{0.25em} & VHD \hspace{0.05em} & All \hspace{0.25em} \\
      \hline
      & MM(KS)	&	1.489	&	0.770	&	1.129	&	\bf 2.922	&	\bf 1.577	\\
      & MM(PP)	&	0.962	&	0.726	&	1.115	&	4.612	&	1.854	\\
      & CC	&	0.817	&	0.774	&	0.601	&	6.961	&	2.288	\\
      & PACC	&	1.012	&	3.326	&	1.234	&	8.283	&	3.464	\\
      & ACC	&	0.831	&	0.696	&	0.505	&	13.726	&	3.940	\\
      & T50	&	0.897	&	0.921	&	0.720	&	15.903	&	4.610	\\
      & X	&	1.051	&	1.035	&	0.733	&	19.178	&	5.499	\\
      & SVM(KLD)	&	\bf 0.601	& \bf 	0.543	&	\bf 0.413	&	22.176	&	5.933	\\
      & MAX	&	4.249	&	4.012	&	1.779	&	15.331	&	6.343	\\
      & PCC	&	15.713	&	10.755	&	36.040	&	93.282	&	38.948	\\
      & MS	&	50.860	&	13.498	&	184.874	&	118.333	&	91.891	\\
      \hline
    \end{tabular}
     }
  \label{tab:raevalues}
\end{table}


\subsection{Testing classification accuracy}
\label{sec:classaccuracy}
 
\noindent As discussed in Section \ref{sec:introduction},
quantification accuracy is related to the classifier's ability to
balance false positives and false negatives, but is \emph{not} related
to its ability to keep their total number low, which is instead a key
requirement in standard classification.  However, it is fairly natural
to expect that a user will trust a quantification method only inasmuch
as its good quantification performance results from reasonable
classification performance. In other words, a user is unlikely to
accept a classifier with good quantification accuracy but bad
classification accuracy.
 

%

For this reason we have compared, in terms of \emph{classification}
accuracy, SVM(KLD) against the traditional classification-oriented
SVMs (i.e., SVM$^{org}$ -- see Section
\ref{sec:novelquantificationmethods}), with the goal of ascertaining
if the former has also a reasonable classification accuracy. Default
parameters have been used for both, for the reasons already explained
in the first paragraph of Section \ref{sec:experiments}. We have
compared the two systems by using the same training set as used in the
quantification experiments, and as the test set the union of the 52
test sets used for quantification (i.e., a single test set of 791,607
documents across 99 classes). Evaluation is based on the $F_{1}$
measure, both in its micro-averaged ($F_{1}^{\mu}$) and macro-averaged
($F_{1}^{M}$) versions.

On the \textsc{RCV1-v2} dataset, in terms of $F_{1}^{\mu}$ SVM(KLD)
performs slightly worse than SVM$^{org}$ (.755 instead of .777, with a
relative decrease of $-2.83\%$), while in terms of $F_{1}^{M}$
SVM(KLD) performs slightly better (.440 instead of .433, a relative
increase of $+1.61\%$), which indicates that, on highly imbalanced
classes, SVM(KLD) is not only a better quantifier, but also a better
classifier.  To witness, on the 48 classes for which $\lambda_{Tr}\leq
0.01$ (i.e., on the most imbalanced classes) we obtain
$F_1^{\mu}=.294$ for SVM$^{org}$ while SVM(KLD) obtains
$F_1^{\mu}=.350$ (an increase of $+19.09\%$). The trends on the
\textsc{OHSUMED-S} dataset are similar, though with smaller margins.
In terms of $F_{1}^{\mu}$ SVM(KLD) performs slightly worse than
SVM$^{org}$ ($F_{1}^{\mu}=.713$ instead of $.722$, with a relative
decrease of $-1.24\%$), while in terms of $F_{1}^{M}$ SVM(KLD)
performs slightly better (.405 instead of .398, a relative increase of
$+1.76\%$), confirming the insights obtained from \textsc{RCV1-v2}.
On the 51 classes for which $\lambda_{Tr}\leq 0.01$ (i.e., on the most
imbalanced classes) we obtain $F_1^{\mu}=.443$ for SVM$^{org}$ while
SVM(KLD) obtains $F_1^{\mu}=.456$ (an increase of $+2.93\%$).

All in all, these results show that classifiers trained via SVM(KLD),
aside from delivering top-notch quantification accuracy, are also
characterized by very good classification accuracy. This makes the
quantifiers generated via SVM(KLD) not only accurate, but also
trustworthy.

\subsection{Testing efficiency}
\label{sec:efficiency}

\noindent SVM(KLD) has also good properties in terms of sheer
efficiency.

Concerning training, Joachims \citeyear{Joachims05} proves that
training a classifier with SVM$_{perf}$ is $O(n^2)$, with $n$ the
number of training examples, for any loss function that can be
computed from a contingency table (such as KLD indeed is). This is
certainly more expensive than training a classifier by means of a
standard, linear SVM (which is at the heart of Forman's implementation
of all the quantification methods of \S
\ref{sec:existingquantificationmethods}), since this latter is
well-known to be $O(sn)$ (with $s$ the average number of non-zero
features in the training objects) \cite{Joachims06}.

However, note that, while SVM(KLD) only requires training a classifier
by means of SVM$_{perf}$, setting up a quantifier with any of the
methods of \S \ref{sec:existingquantificationmethods} (with the only
exception of the simple CC method) requires more than simply training
a classifier. For instance, ACC (together with the methods derived
from it, such as T50, X, MAX, MS, PACC) also requires estimating
$tpr_{Te}$ and $fpr_{Te}$ on the training set via $k$-fold cross
validation, which may be expensive; analogously, both MM(KS) and
MM(PP) require estimating $D_{c}^{Te}$ and $D_{\overline{c}}^{Te}$ via
$k$-fold cross-validation, and the same considerations apply.

In practice, using SVM$_{perf}$ turns out to be affordable. On
\textsc{RCV1-v2}, training the 99 binary classifiers described in the
previous sections via SVM$_{perf}$ required on average about 4.7
seconds each\footnote{All times reported in this section were measured
on a commodity machine equipped with an Intel Centrino Duo
2$\times$2Ghz processor and 2GB RAM.}. By contrast, training the
analogous classifiers via a standard linear SVM required on average
only 2.1 seconds each. However, this means that, if $k$-fold
cross-validation is used for the estimation of parameters with a value
of $k\geq 2$ (meaning that, for each class, additional $k$ classifiers
need to be trained), the computational advantage of using a linear SVM
instead of the more expensive SVM$_{perf}$ is completely lost. Forman
\citeyear{Forman:2008kx} recommends choosing $k=50$ in order to obtain
more accurate estimates of $tpr_{Te}$ and $fpr_{Te}$ for use in ACC
and derived methods; this means making the training phase roughly
$(2.1\cdot 51)/4.7\approx 22$ times slower than the training phase of
SVM(KLD).

Concerning the computational costs involved at classification /
quantification time, SVM(KLD) and all the baseline methods discussed
in this paper are equivalent, since (a) they all generate linear
classifiers of equal efficiency, and (b) in ACC and derived methods
the cost of the post-processing involved in computing class
prevalences from the classification decisions is negligible.

%
%


\section{Related work}
\label{sec:related}

\noindent An early mention of quantification can be found in
\cite[Section 7]{Lewis95}, where this task is simply called
\emph{counting}; however, the author does not propose any specific
solution for this problem. Forman
\citeyear{Forman:2005fk,Forman:2006uq,Forman06,Forman:2008kx} is to be
credited for bringing the problem of quantification to the attention
of the data mining and machine learning research communities, and for
proposing several solutions for performing quantification and for
evaluating it.


\subsection{Applications of quantification}


\noindent Chan and Ng \citeyear{Chan2005,Chan2006} apply
quantification (which they call ``class prior estimation'') to
determining the prevalence of different senses of a word in a text
corpus, with the goal of improving the accuracy of word sense
disambiguation algorithms as applied on that corpus. Forman
\citeyear{Forman:2008kx} uses quantification in order to establish the
prevalence of various support-related issues in incoming telephone
calls received at customer support desks. Esuli and Sebastiani
\citeyear{Esuli:2010kx} apply quantification methods for estimating
the prevalence of various response classes in open-ended answers
obtained in the context of market research surveys (they do not use
the term ``quantification'', and rather speak of ``measuring
classification accuracy at the aggregate level''). Hopkins and King
\citeyear{Hopkins:2010fk} classify blog posts with the aim of
estimating the prevalence of different political candidates in
bloggers' preferences.
Gonzalez-Castro et al.\
\citeyear{Gonzalez-Castro:2013fk} and Sanchez et al.\
\citeyear{Sanchez:2008uq} use quantification for establishing the
prevalence of damaged sperm cells in a given sample for veterinary
applications.  Baccianella et al.\ \citeyear{Baccianella:2013fg}
classify radiology reports with the aim of estimating the prevalence
of different pathologies.  Tang et al.\ \citeyear{Tang:2010uq} focus
on \emph{network} quantification problems, i.e., problems in which the
goal is to estimate class prevalence among a population of nodes in a
network. Alaiz-Rodriguez et al.\ \citeyear{Alaiz-Rodriguez:2011fk},
Limsetto and Waiyamai \citeyear{Limsetto:2011fk}, Xue and Weiss
\citeyear{Xue:2009uq}, and Zhang and Zhou \citeyear{Zhang:2010kx} use
quantification in order to improve classification, i.e., attempt to
estimate class prevalence in the test set in order to generate a
classifier that better copes with differences in the class
distributions of the training set and the test set.

Many other works use quantification ``without knowingly doing so'';
that is, unaware of the existence of methods specifically optimized
for quantification, they use classification with the only goal of
estimating class prevalences. In other words, these works use plain
``classify and count''. Among them, Mandel et al.\
\citeyear{Mandel:2012dq} use tweet quantification in order to
estimate, from a quantitative point of view, the emotional responses
of the population (segmented by location and gender) to a natural
disaster; O'Connor et al.\ \citeyear{OConnor:2010fk} analyse the
correlation between public opinion as measured via tweet sentiment
quantification and via traditional opinion polls; Dodds et al.\
\citeyear{Dodds:2011uq} use tweet sentiment quantification in order to
infer spatio-temporal happiness patterns; and Weiss et al.\
\citeyear{Weiss:2013fk} use quantification in order to measure the
prevalence of different types of pets' activity as detected by
wearable devices.

%


\subsection{Quantification methods}

\noindent Bella et al.\ \citeyear{Bella:2010kx} compare many of the
methods discussed in \S \ref{sec:existingquantificationmethods}, and
find that CC $\prec$ PCC $\prec$ ACC $\prec$ PACC (where $\prec$ means
``underperforms''). Also Tang et al.\ \citeyear{Tang:2010uq}
experimentally compare several among the methods discussed in \S
\ref{sec:existingquantificationmethods}, and find that CC $\prec$ PCC
$\prec$ ACC $\prec$ PACC $\prec$ MS. They also propose a method
(specific to linked data) that does not require the classification of
individual items, but they find that it underperforms a robust
classification-based quantification method such as MS.  However, the
experimental comparisons of \cite{Bella:2010kx} and \cite{Tang:2010uq}
are both framed in terms of absolute error, which seems a sub-standard
evaluation measure for this task (see \S
\ref{sec:evaluationmeasures}); additionally, the datasets they test on
do not exhibit the severe imbalance typical of many binary text
classification tasks, so it is not surprising that their results concerning
MS are not confirmed by our experiments.

The idea of using a learner that directly optimizes a loss function
specific to quantification was first proposed, although not
implemented, in \cite{Esuli:2010fk}, which indeed proposes using
SVM$_{perf}$ to directly optimize KLD; the present paper is thus the
direct realization of that proposal. The first published work that
implements and tests the idea of directly optimizing a
quantification-specific loss function is \cite{Milli:2013fk}, whose
authors propose variants of decision trees and decision forests that
directly optimize a loss combining classification accuracy and
quantification accuracy. At the time of going to print we have become
aware of a related paper \cite{Barranquero:2015fr} whose authors,
following \cite{Esuli:2010fk}, use $SVM_{perf}$ to perform
quantification; differently from the present paper, and similarly to
\cite{Milli:2013fk}, they use an evaluation function that combines
classification accuracy and quantification accuracy.


%







\subsection{Other related work}

\noindent Bella et al.\ \citeyear{Bella:2014kp} address the problem of
performing quantification for the case in which the output variable to
be predicted for a given individual is a real value, instead of a
class as in the case we analyse; that is, they analyse quantification
as a counterpart of \emph{regression}, rather than of classification
as we do.

Quantification as defined in this paper bears some relation with
density estimation \cite{Silverman:1986fk}, which can be defined as
the task of estimating, based on observed data, the unknown
probability density function of a given random variable. If the random
variable is discrete, this means estimating, based on observed data,
the unknown distribution across the discrete set of events, i.e.,
across the classes. An example density estimation problem is
estimating the percentage of white balls in a very large urn
containing white balls and black balls. However, there are two
essential differences between quantification and density estimation,
i.e., that (a) in density estimation the class to which a data item
belongs can be established with certainty (e.g., if a ball is picked,
it can be decided with certainty if it is black or white), while in
quantification this is not true; and (b) in density estimation the
population of data items is usually so large as to make it infeasible
to check the class to which each data item belongs (e.g., only a
limited sample of balls is picked), while this is not true in
quantification, where it is assumed that all the items can be checked
for the purpose of estimating the class distribution.

A research area that might seem related to quantification is
\emph{collective classification} (CoC) \cite{Sen:2008fk}. Similarly to
quantification, in CoC the classification of instances is not viewed
in isolation. However, CoC is radically different from quantification
in that its focus is on improving the accuracy of classification by
exploiting relationships between the objects to classify (e.g.,
hypertextual documents that link to each other). Differently from
quantification, CoC (a) assumes the existence of explicit
relationships between the objects to classify, which quantification
does not, and (b) is evaluated at the individual level, rather than at
the aggregate level as quantification.

Quantification bears strong relations with \emph{prevalence estimation
from screening tests}, an important task in epidemiology (see
\cite{Levy:1970fk,Lew:1989fk,Kuchenhoff2012,Rahme:1998uq,Zhou:2002fk}). A
screening test is a test that a patient undergoes in order to check if
s/he has a given pathology. Tests are often imperfect, i.e., they may
give rise to false positives (the patient is incorrectly diagnosed
with the pathology) and false negatives (the test wrongly diagnoses
the patient to be free from the pathology). Therefore, testing a
patient is akin to classifying a document, and using these tests for
estimating the prevalence of the pathology in a given population is
akin to performing aggregative quantification. The main
difference between this task and quantification is that a screening
test typically has known and fairly constant recall (that
epidemiologists call ``sensitivity'') and fallout (whose complement
epidemiologists call ``specificity''), while the same usually does not
happen for a classifier.




\section{Conclusions}
\label{sec:conclusions}

\noindent We have presented SVM(KLD), a new method for performing
quantification, an important (if scarcely investigated) task in
supervised learning, where estimating class prevalence, rather that
classifying individual items, is the goal. The method is sharply
different from most other methods presented in the literature. While
most such methods adopt a general-purpose classifier (where the
decision threshold has possibly been tuned according to some
heuristics) and adjust the outcome of the ``classify and count''
phase, we adopt a straightforward ``classify and count'' approach
(with no threshold tuning and/or \textit{a posteriori} adjustment) but
generate a classifier that is directly optimized for the evaluation
measure used for estimating quantification accuracy. This is not
straightforward, since an evaluation measure for quantification is
inherently non-linear and multivariate, and thus does not lend itself
to optimization via standard supervised learning algorithms. We
circumvent this problem by adopting a supervised learning method for
structured prediction that allows the optimization of non-linear,
multivariate loss functions, and extend it to optimize KLD, the
standard evaluation measure of the quantification literature.

Experimental results that we have obtained by comparing SVM(KLD) with
ten different state-of-the-art
baselines 
show that SVM(KLD) (i) is more accurate (in a statistically
significant way) than the competition, (ii) is more stable than the
tested baselines, since it systematically shows very good performance
irrespectively of class prevalence (i.e., level of imbalance) and
distribution drift (i.e., discrepancy between the class distributions
in the training and in the test set), (iii) is also accurate at the
classification (aside from the quantification) level, and (iv) is 20
times faster to train than the competition. These experiments have
been run, against 10 state-of-the-art baseline methods, on a batch of
5,148 binary, high-dimensional datasets (averaging more than 15,200
documents each and characterized by more than 50,000 features) and on
a further batch of 352 binary, high-dimensional datasets (averaging
more than 3,200 documents each and characterized by more than 10,000
features), all characterized by varying levels of class imbalance and
distribution drift. These figures qualify the present experimentation
as a very large one.

\begin{acks}
We are indebted to George Forman for letting us have the
code for all the quantification methods he introduced in
\cite{Forman:2005fk,Forman06,Forman:2008kx}, which we have used as
baselines.
We are also very grateful to Thorsten Joachims for making his
SVM$_{perf}$ package available, and for useful discussions about its
use in quantification.
\end{acks}


\newpage

\appendix

\section{Appendix: On the presumed invariance of $tpr$ and $fpr$
across test sets}
\label{sec:ratioStudy}

\noindent Most methods described in \S
\ref{sec:existingquantificationmethods} (specifically: ACC, PACC, T50,
X, MAX, MS) rely, among other things, on the assumption that
$tpr_{Te}$ and $fpr_{Te}$ can reliably be estimated via $k$-fold
cross-validation on the training set; in other words, they rely on the
assumption that $tpr$ and $fpr$ do not change when the classifier is
applied to different test sets.

Indeed, Forman explicitly assumes that the application domains he
confronts are of a type that \cite{Fawcett:2005fk} call
``$y\rightarrow \textbf{x}$ domains'', in which $tpr$ and $fpr$ are in
fact invariant with respect to the test set the classifier is applied
to. The notation $y\rightarrow \textbf{x}$ means that the value of the
$y$ variable (the output label) probabilistically determines the
values of the $\textbf{x}$ variables (the input features), i.e.,
$\textbf{x}$ causally depends on $y$; in other words, the
class-conditional probabilities $p(\textbf{x}|y)$ are invariant across
different test sets. For instance, our classification problem may
consist in predicting whether a patient suffers or not from a given
pathology (a fact represented by a binary variable $y$) given a vector
$\textbf{x}$ of observed symptoms. This is indeed a ``$y\rightarrow
\textbf{x}$ domain'', since the causality relation is from $y$ to
$\textbf{x}$, i.e., it is the presence of the pathology that
determines the presence of the symptoms, and not vice versa. In other
words, the class conditional probabilities $p(\textbf{x}|y)$ do not
change across test sets: if more people suffer from the pathology,
more people will exhibit its symptoms. Important quantification
problems within $y\rightarrow \textbf{x}$ domains do exist. One of
them is when epidemiologists attempt to estimate the prevalence of
various causes of death ($y$) from ``verbal autopsies'', i.e., binary
vectors of symptoms (\textbf{x}) as extracted from oral accounts
obtained from relatives of the deceased \cite{King:2008fk}.


However, many application domains are instead of the type that
\cite{Fawcett:2005fk} call ``$\textbf{x} \rightarrow y$ domains'',
where it is $y$ that causally depends on $\textbf{x}$, and not vice
versa. For instance, our classification problem may consist in
predicting whether the home football team is going to win tomorrow's
match or not ($y$) given a number of stats ($\textbf{x}$) about its
recent performance (e.g., number of goals scored in the last $k$
matches, number of goals conceded, etc.). This is indeed a
``$\textbf{x} \rightarrow y$ domain'', since the (assumed) causality
relation is from $\textbf{x}$ to $y$, i.e., the recent performance of
the team is an indicator of its state of form, which may
probabilistically determine the outcome of the game; it is certainly
\emph{not} the case that the outcome of the game determines the past
performance of the team! And in $\textbf{x} \rightarrow y$ domains the
class-conditional probabilities $p(\textbf{x}|y)$ are not guaranteed
to be constant.

One may wonder whether \emph{text} classification contexts are
$y\rightarrow \textbf{x}$ or $\textbf{x} \rightarrow y$ domains. Given
the wide array of uses text classification has been put to, we think
it is difficult to make general statements about this. We prefer to
follow the advice in \cite{Fawcett:2005fk}, who recommend ``that
researchers test their assumptions in practice''. We have thus
compared, for each of our 5,148 \textsc{RCV1-v2} test sets and 352
\textsc{OHSUMED-S} test sets, (a) the $tpr_{Te}$ and $fpr_{Te}$ values
that the standard linear SVM classifier (the one at the heart of all
our baseline methods) has obtained, with (b) the corresponding
$tpr_{Tr}$ and $fpr_{Tr}$ values that we have computed on the training
set via $k$-fold cross validation. If $tpr$ and $fpr$ were invariant
across different sets, then (a) and (b) should be approximately the
same. The results, which are displayed in Table
\ref{tab:trainTestRatioDifferences}, show that in our domain $tpr$ and
$fpr$ are far from being invariant across different sets. For
instance, on \textsc{RCV1-v2} the average $tpr_{Tr}$ as computed via
$k$-fold cross-validation is 0.443, while the analogous average on the
5,148 test sets is 0.357, a -19.29\% decrease; interestingly enough,
on \textsc{OHSUMED-S} we witness an analogous trend but with opposite
sign, with a +59.48\% variation (from 0.196 to 0.313) in going from
training to test sets\footnote{Note that the average pairwise
difference between two sets of values is \emph{greater or} equal than
the difference of their averages; thus, the discrepancy between the
training set values and the corresponding test set values may actually
be even higher than Table \ref{tab:trainTestRatioDifferences}
shows.}. A similar although less marked pattern can be observed for
$fpr$.

In conclusion, $tpr$ and $fpr$ turn out to be far from being invariant
across different sets, at least in the application contexts from which
our datasets are drawn. It is thus evident that, at the very least in
text quantification contexts, assuming that $tpr$ and $fpr$ are indeed
invariant, and adopting methods that rely on this assumption, is
risky, and definitely suboptimal in some cases. This is yet another
reason to prefer methods, such as SVM(KLD), which do not rely on any
such assumption.

\begin{table}[t]
  \tbl{Average $tpr$ and $fpr$ values measured on the training and test sets of the \textsc{RCV1-v2} and \textsc{OHSUMED-S} datasets. The values in the avg($tpr_{Tr}$) and avg($fpr_{Tr}$) are computed via $k$-fold cross-validation, and are thus averages across 99 classes (\textsc{RCV1-v2}) and 88 classes (\textsc{OHSUMED-S}). The values in the avg($tpr_{Te}$) and avg($fpr_{Te}$) are averages across 5,148 test sets (\textsc{RCV1-v2}) and 352 test sets (\textsc{OHSUMED-S}). Column 4 and 7 show the relative variation in these values, measured as $(tpr_{Te}-tpr_{Tr})/(tpr_{Tr})$ and $(fpr_{Te}-fpr_{Tr})/(fpr_{Tr})$, respectively.}{
  \begin{tabular}{|r||ccr|ccr|}
    \hline
    &  avg($tpr_{Tr}$) &  avg($tpr_{Te}$) & rel \% diff &  avg($fpr_{Tr}$) &  avg($fpr_{Te}$)  & rel \% diff \\\hline
    \textsc{RCV1-v2} & 0.443 & 0.357 & -19.29\% & 2.68E-03 & 2.54E-03 & -5.12\% \\
    \textsc{OHSUMED-S} & 0.196 & 0.313 & +59.48\% & 1.41E-03 & 1.82E-03 & +29.09\% \\
    \hline
  \end{tabular}
  }
  \label{tab:trainTestRatioDifferences}
\end{table}

\bibliographystyle{ACM-Reference-Format-Journals} \bibliography{New-TKDD2015(final)}

\end{document}